\documentclass{article} 
\usepackage{iclr2026_conference,times}

\usepackage{amsmath,amsfonts,bm}

\def\eqref#1{equation~\ref{#1}}

\def\1{\bm{1}}

\DeclareMathAlphabet{\mathsfit}{\encodingdefault}{\sfdefault}{m}{sl}
\SetMathAlphabet{\mathsfit}{bold}{\encodingdefault}{\sfdefault}{bx}{n}

\usepackage{hyperref}
\usepackage{url}

\newcommand{\OursMethod}{EditVerse}
\newcommand{\OursBench}{EditVerseBench}

\usepackage{multirow} 
\usepackage{graphicx}
\usepackage{booktabs}
\usepackage[table]{xcolor}
\usepackage{makecell}
\usepackage{threeparttable}
\usepackage{wrapfig}
\usepackage{pifont}
\usepackage{caption}
\usepackage{hyperref}
\usepackage{subfigure}
\usepackage{subcaption}

\definecolor{dark_red}{HTML}{e10000}
\definecolor{dark_blue}{HTML}{0075EE}

\definecolor{lightgray}{rgb}{0.8, 0.8, 0.8}

\newcommand{\graymidrule}{\arrayrulecolor{lightgray}\midrule\arrayrulecolor{black}}
\newcommand{\graybottomrule}{\arrayrulecolor{lightgray}\bottomrule\arrayrulecolor{black}}

\iclrfinalcopy

\title{EditVerse: Unifying Image and Video Editing and Generation with In-Context Learning}

\author{Xuan Ju$^{1,2}$\thanks{This work was done when Xuan Ju was an intern at Adobe Research.}\quad
    Tianyu Wang$^{1}$\quad
    Yuqian Zhou$^{1}$\quad
    He Zhang$^{1}$\quad
    Qing Liu$^{1}$\quad
    Nanxuan Zhao$^{1}$\\
    \textbf{Zhifei Zhang}$^{1}$\quad
    \textbf{Yijun Li}$^{1}$\quad
    \textbf{Yuanhao Cai}$^{3}$\quad
    \textbf{Shaoteng Liu}$^{1}$\quad
    \textbf{Daniil Pakhomov}$^{1}$\quad
    \textbf{Zhe Lin}$^{1}$\\
    \textbf{Soo Ye Kim}$^{1}$\thanks{Corresponding author.}\quad
    \textbf{Qiang Xu}$^{2,4\dag}$ \\[1em] 
    $^1$Adobe Research \quad
    $^2$CUHK \quad
    $^3$Johns Hopkins University \quad
    $^4$Shenzhen Loop Area Institute
}

\begin{document}

\maketitle

\begin{center}
\vspace{-12mm}
    \centering
    \includegraphics[width=0.98\textwidth]{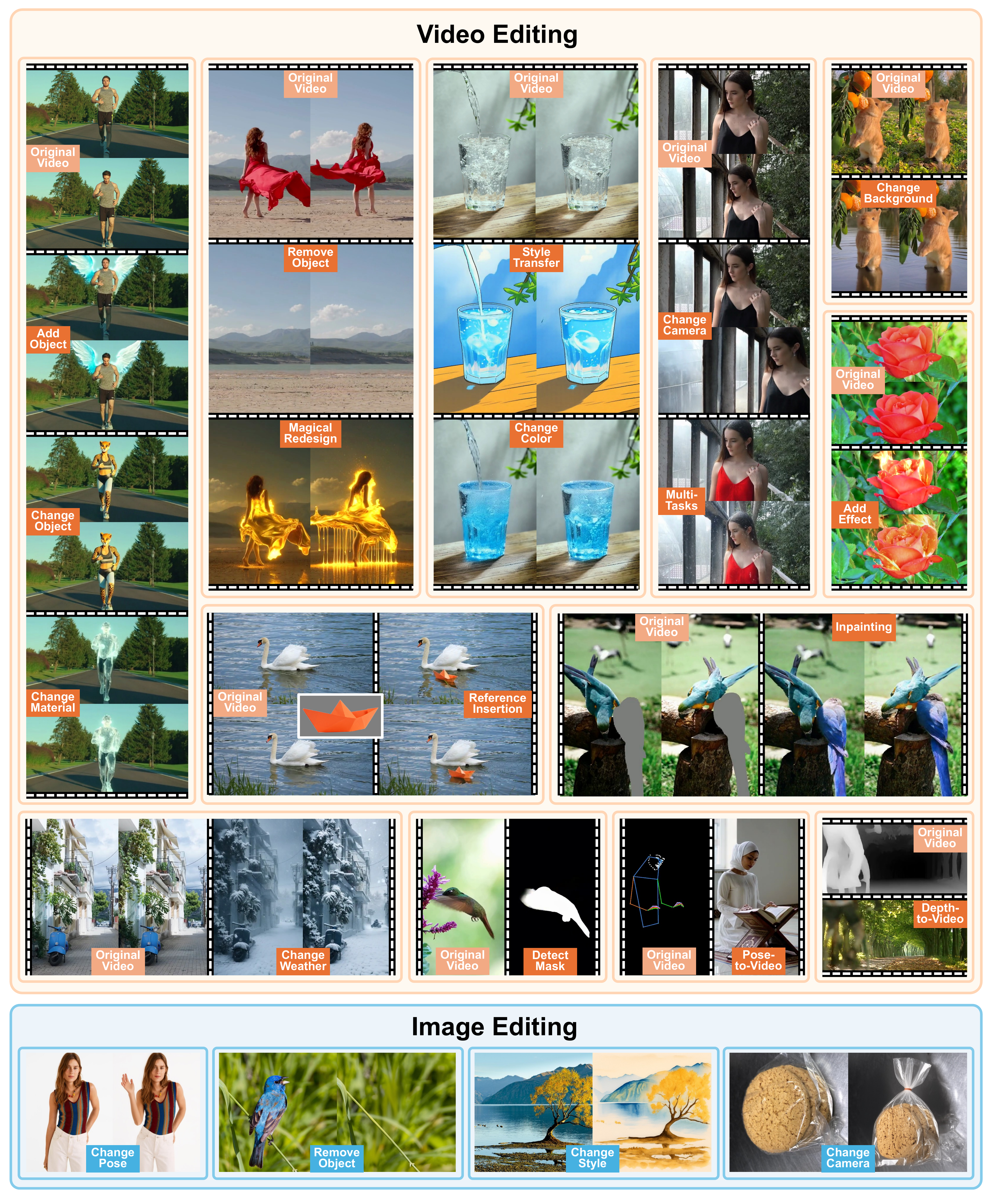}
\vspace{-2mm}
    \captionsetup{font={stretch=0.8}}
    \captionof{figure}{The strong video editing performance of \OursMethod~\textbf{emerges from a unified framework} trained on a diverse set of mixed image and video data. This teaser visualizes a selection of supported image and video editing tasks (Instructions in the Appendix). More results in our \href{http://editverse.s3-website-us-east-1.amazonaws.com/}{\textit{Project Page}}.
    }
    \label{fig:teaser}
\end{center}

\begin{abstract}

Recent advances in foundation models highlight a clear trend toward unification and scaling, showing emergent capabilities across diverse domains. While image generation and editing have rapidly transitioned from task-specific to unified frameworks, video generation and editing remain fragmented due to architectural limitations and data scarcity. In this work, we introduce \OursMethod, a unified framework for image and video generation and editing within a single model. By representing all modalities, \textit{i.e.}, text, image, and video, as a unified token sequence, \OursMethod~leverages self-attention to achieve robust in-context learning, natural cross-modal knowledge transfer, and flexible handling of inputs and outputs with arbitrary resolutions and durations. To address the lack of video editing training data, we design a scalable data pipeline that curates 232K video editing samples and combines them with large-scale image and video datasets for joint training. Furthermore, we present \OursBench, the first benchmark for instruction-based video editing covering diverse tasks and resolutions. Extensive experiments and user studies demonstrate that \OursMethod~achieves state-of-the-art performance, surpassing existing open-source and commercial models, while exhibiting emergent editing and generation abilities across modalities.
\end{abstract}

\vspace{-4mm}
\section{Introduction}
\label{sec:introduction}
\vspace{-2mm}

Recent advancements of foundation models in computer vision and large language models highlight a clear trend toward unification and scaling ~\citep{gpt4,transfusion,begal}, showing that joint training on diverse datasets can unlock emergent intelligence.
Specifically in image generation and editing, there is also a shift from domain-specific models~\citep{controlnet,humansd,photomaker} toward universal models~\citep{kontext,unireal} that unify diverse generation and editing tasks under a generalized and scalable framework.

However, unlike the image domain, the exploration of unified video generation and editing remains limited~\citep{vace,unic}.
This stems from two primary challenges: 
\textbf{(1) Architectural Limitations}: Existing video generation models, mostly based on cross-attention~\citep{moviegen,wanx} or MMDiT~\citep{cogvideox,hunyuanvideo} architecture, are typically designed for specific tasks such as text-to-video generation. Adapting them to support various video generation and editing tasks introduces substantial design and scaling challenges. 
For example, VACE~\citep{vace} uses an additional branch that accepts unedited videos and masks as input, transforming a text-to-video model into a video inpainting model. 
However, it relies on masks to localize the editing regions and requires task-specific input configurations, making it less practical for real-world use.
To unlock emergent abilities with in-context learning, a fully unified framework must be able to process diverse input modalities (e.g., text, image, video) and types (e.g., duration, resolution) with a consistent and flexible representation.
\textbf{(2) Data Scarcity and Diversity}: Unlike the vast and varied datasets readily available for image editing~\citep{anyedit,imgedit,sharegpt4oimg}, high-quality and diverse video editing datasets are significantly scarce.

To address this challenge, we propose \textbf{\OursMethod}, a unified framework that enables image and video editing and generation within a single model, leveraging full self-attention to enable robust in-context learning and effective knowledge transfer between images and videos.
Our design considers two aspects:
\textbf{(1) In-Context Learning}: We represent all modalities (text, image, and video) as a unified one-dimensional token sequence, which is then concatenated and fed into the model as a long sequence.
This design enables the use of full self-attention with strong in-context learning capabilities~\citep{fulldit} to jointly model and align different modalities.
As a result, \OursMethod~achieves enhanced text comprehension, improved image and video editing quality, and most importantly, natural cross-modal knowledge transfer between images and videos, which effectively alleviates the limitations caused by the scarcity of video editing data.
\textbf{(2) Flexibility}: We use an interleaved design for text, image, and video, inspired by the native generation architecture of multimodal large language models (MLLM), which are well-suited for supporting diverse tasks and interactive generation. This design enables the model to process image and video inputs and outputs with arbitrary resolution, temporal duration, and sequential position, thereby providing enhanced flexibility. To further distinguish positional and modal information, we introduce a four-dimensional Rotary Positional Embedding (RoPE) that incorporates sequential, temporal, height, and width dimensions.

While careful model design is crucial, simply training it on image editing data is insufficient to enable the model to perform various video editing tasks.
Based on the observation that open-source instruction-based video datasets~\citep{senorita-2m} are inadequate in both volume and quality, we devise a data pipeline that first generates video editing samples with task-specific models, then filters high-quality samples from the generated samples. 
For our unified training, we mix such curated video editing data ($232K$) with $56K$ samples filtered from Señorita-2M as well as $2M$ image generation samples, $6M$ image editing samples, and $4M$ video generation samples.

At last, due to the absence of instruction-based video editing benchmarks encompassing diverse tasks and mixed resolutions, we introduce \textbf{\OursBench}~to enable a more comprehensive evaluation. It contains $100$ videos, evenly divided between $50$ horizontal and $50$ vertical formats, with each video paired with two editing prompts in different editing tasks. Each data instance includes an editing instruction, a source prompt, and a target prompt, spanning $20$ distinct video editing categories. Comprehensive evaluations (both automated and user studies) demonstrate that \OursMethod~achieves state-of-the-art performance compared to existing open-source methods as well as commercial models. Moreover, experiment results show the model’s capacity for knowledge transfer from image to video domain and reveal emergent abilities arising from our proposed design.

\vspace{-6mm}
\section{Related Work}
\label{sec:related_work}
\vspace{-3mm}

\noindent \textbf{Instruction-based Image and Video Editing Datasets.} 
In recent years, the field has witnessed a surge in large-scale, open-source datasets for instruction-based image editing. 
Increasingly sophisticated data annotation pipelines have been designed and continuously improved, advancing from earlier methods doing large-scale annotation using editing models with lower success rates (\textit{e.g.}, InstructPix2Pix~\citep{instructpix2pix} and HQ-Edit~\citep{hqedit}) or small-scale manual labeling (\textit{e.g.}, MagicBrush~\citep{magicbrush}), to advanced techniques that leverage well-trained task-specific models and pipelines to generate better quality data at a large scale (\textit{e.g.}, UltraEdit~\citep{ultraedit}, OmniEdit~\citep{omniedit}, AnyEdit~\citep{anyedit},  SEED-Data-Edit~\citep{seed-data-edit}, and EditWorld~\citep{editworld}).
Data quality improvement further boosts the performance of instruction-based image editing models~\citep{gpt-4o,kontext}, which are then served as a data source of open-source datasets (\textit{e.g.}, ShareGPT-4o-Image~\citep{sharegpt4oimg}).

However, video editing datasets progress at a slower pace. 
InsV2V~\citep{insv2v} uses Prompt-to-Prompt~\citep{prompt-to-prompt} and a large language model (LLM) to create its video editing datasets, where the low performance ceiling of Prompt-to-Prompt leads to poor dataset quality.
Although VIVID-10M~\citep{vivid-10m} provides a collection of videos with corresponding textual instructions and mask annotations, it lacks paired ground-truth edited videos, making it unsuitable for training instruction-based video editing models. 
Señorita-2M~\citep{senorita-2m} builds an instruction-based video editing dataset using task-specific diffusion models. However, when compared to datasets in image editing, it exhibits notable limitations in both quality and editing diversity. 
In conclusion, datasets for instruction-based video editing are significantly less mature than the image domain, necessitating architectural innovation to transfer editing capabilities from image to video.

\noindent \textbf{Image and Video Editing.}
The success of diffusion models has led to rapid progress in image and video editing. 
Since most pre-trained models are designed for text-to-image and text-to-video generation, early research explores training-free image and video editing techniques based on these models, often by manipulating attention maps or latent spaces~\citep{prompt-to-prompt,masactrl,directinversion,vid2vidzero,fatezero,videop2p,raccoon}. 
Despite their simplicity, such techniques frequently yield unsatisfactory results characterized by a lack of precise control and low quality.
Consequently, the field has shifted mainly to data-driven training-based methods. 
For image editing, methods such as InstructPix2Pix~\citep{instructpix2pix} and subsequent works~\citep{emuedit} concatenate the unedited image latent with the noisy latent along the channel dimension, directly fine-tuning text-to-image models for editing tasks. 
Later studies~\citep{fulldit,omnigen,hidreami1technicalreport,icedit} find that sequential concatenation benefits model learning by using self-attention to improve in-context learning, which is a design choice that also aligns with the architectures of multimodal LLMs' native image generation (\textit{e.g.}, BAGEL~\citep{begal}, transfusion~\citep{transfusion}). 
While similar techniques can be employed for video editing, investigations into instruction-based video editing are relatively rare.
EVE~\citep{eve} trains adapters on top of frozen text-to-image models to enable video editing ability.
InsV2V~\citep{insv2v} extends InstructPix2Pix~\citep{instructpix2pix} to a video version.
GenProp~\citep{genprop} propagates the edits in the given first frame to the following frames.
Recent work UNIC~\citep{unic} concatenates conditions sequentially, similar to image editing architecture designs, and supports six editing tasks with task-aware positional embeddings.
However, these methods still fall short in supporting flexible instruction-based video editing tasks.

\vspace{-2mm}
\section{Method}
\label{sec:method}
\vspace{-2mm}

\begin{figure*}[tbph]
    \vspace{-2mm}
    \centering
    \includegraphics[width=0.99\linewidth]{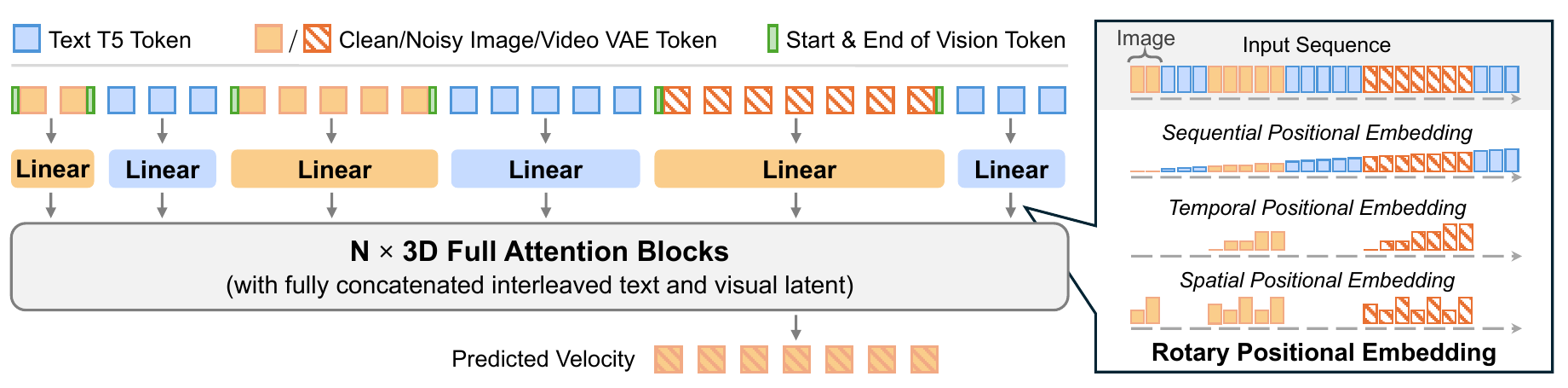}
    \vspace{-2mm}
    \caption{\textbf{Overview of \OursMethod}. We design a unified framework for image and video editing and generation, which processes text and vision inputs into a unified sequence. 
    The right part of the figure shows our positional embedding design.
    This framework leverages full self-attention to facilitate robust in-context learning and effective knowledge transfer among modalities.
    } 
    \vspace{-3mm}
\label{fig:model}
\end{figure*}

As illustrated in Figure~\ref{fig:model}, \OursMethod~employs a transformer architecture with full self-attention~\citep{unireal,fulldit}. 
All text and vision inputs are tokenized and concatenated into a unified sequence in an interleaved manner, then fed into the model and processed with self-attention blocks (Section~\ref{sec:interleaved_text_and_vision_input}). 
To accommodate this interleaved design, we design a four-dimensional Rotary Positional Embedding with spatial (height and width), sequential, and temporal dimensions (Section~\ref{sec:rotary_positional_embedding}).
For training and inference, \OursMethod~predicts the visual velocity~\citep{sd3,flowmatching} that guides the generation of images or videos through a denoising procedure (Section~\ref{sec:training_and_inference_paradigm}).
Subsequent sections detail our framework and the insights behind our design choices.

\vspace{-2mm}

\subsection{Interleaved Text and Vision Input}
\label{sec:interleaved_text_and_vision_input}
\vspace{-2mm}

Following prior works~\citep{vae}, we encode the RGB pixel-space videos and images into a learned spatio-temporally compressed latent space by training a convolutional Variational Autoencoder (VAE) capable of both feature extraction and reconstruction. 
Specifically, given an input image or video $I_{vision}$, the VAE compresses it into a continuous-valued latent representation with downsampling ratios $r_T$, $r_H$, $r_W$. 
Then, the vision features are patchified into a long token sequence with a $1\times 2\times 2$ kernel to get $X_{vision} \in \mathbb{R}^{L_{vision}\times C_{vision}}$ ($L_{vision}$ is the vision token number, $C_{vision}$ is the channel dimension of vision feature).
For a given text input $I_{text}$, we first generate text tokens using the Flan-T5-XXL~\citep{flant5xxl} encoder. 
Then, we retain only the tokens that correspond directly to the input text, discarding the rest, yielding a final representation $X_{text}\in \mathbb{R}^{L_{text}\times C_{text}}$ ($L_{text}$ is the token count of $I_{text}$, $C_{text}$ is the channel dimension of Flan-T5-XXL), which saves computation while preserving the necessary information from text input.

\begin{figure*}[htbp]
    \centering
    \includegraphics[width=0.98\linewidth]{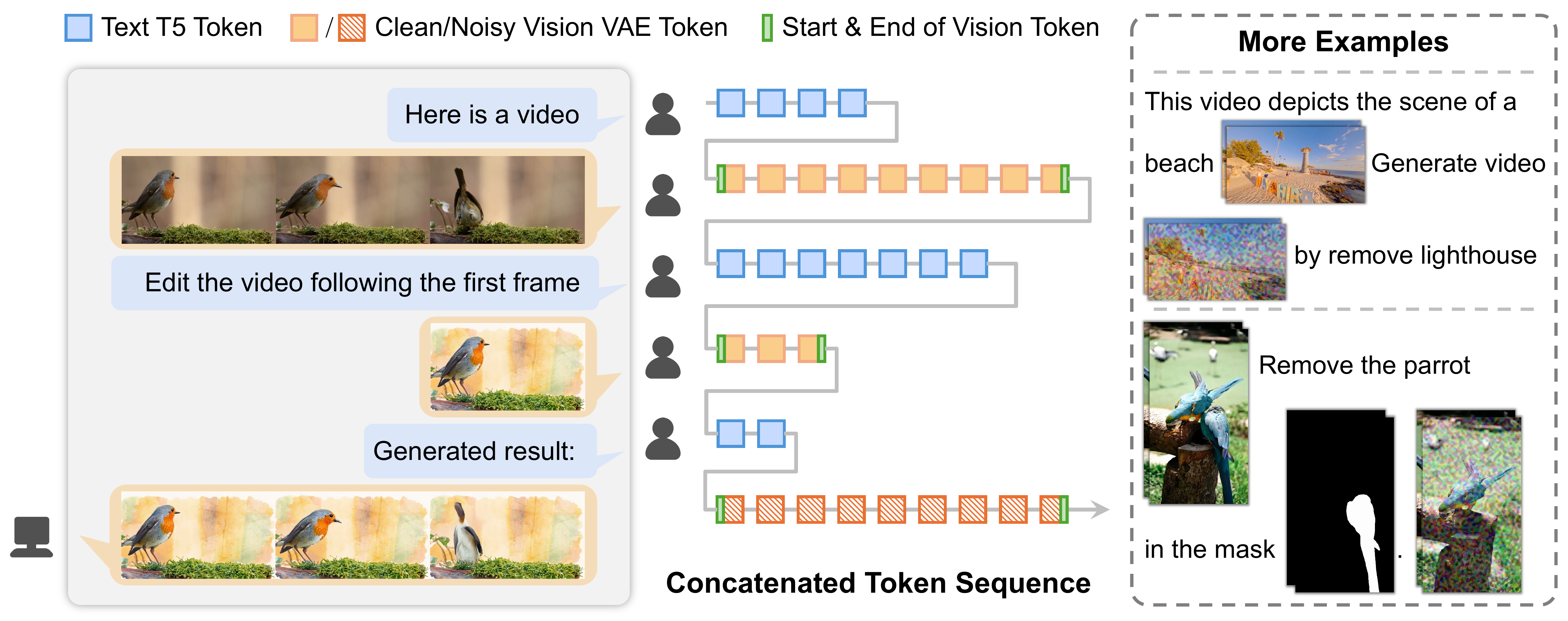}
    \caption{\textbf{Examples for the interleaved text and vision pattern}. \OursMethod~is capable of processing image and video inputs and outputs of arbitrary resolution, duration, and sequential positions.
    } 
\label{fig:interleave}
\end{figure*}

To handle instructions composed of arbitrary combinations of text, images, and videos, we unify all modalities into a single interleaved sequence representation (shown in Figure~\ref{fig:interleave}). 
First, we project the tokens from each modality into a shared embedding space using separate single-layer linear projectors. This maps both text and visual inputs to the model's hidden dimension, $C$, yielding two distinct embedding matrices: $X_{text\_align}\in \mathbb{R}^{L_{text}\times C}$ and $X_{vision\_align}\in \mathbb{R}^{L_{vision}\times C}$. 
Then, we concatenate the projected embeddings to construct a unified input sequence, $X \in \mathbb{R}^{L \times C}$, where $L$ denotes the total number of text and vision tokens. The sequence preserves the original interleaved order of text and visual elements from the instruction.
To explicitly indicate the location of vision tokens (images and videos) within an interleaved sequence, we add a learnable ``start of vision" token and a learnable ``end of vision" token at the beginning and the end of each vision token segment.

\subsection{Rotary Positional Embedding}
\label{sec:rotary_positional_embedding}

To distinguish text, image, and video from each other and to indicate their sequential positions, we design a special Rotary Positional Embedding (RoPE) that incorporates sequential, temporal and spatial (height and width) dimensions (shown in Figure~\ref{fig:model}). 
For each of these four positional dimensions, we apply a separate RoPE computation. 
(1) Sequential dimension: This dimension captures the global position within the overall sequence, starting from $0$. The value is incremented by $1$ for each text token and image/video frame, up to the end of the sequence.
(2) Temporal dimension: This dimension is used exclusively for video frames to encode their temporal order within a video clip. It begins at $0$ and increases by $1$ for each subsequent frame. For text and image inputs, this dimension remains $0$.
(3) Height and Width Dimensions: For images and video frames, the height and width dimensions correspond to the pixel coordinates, increasing incrementally from the top-left to the bottom-right corner~\citep{moviegen}. The increment values reflect the number of pixels along the height and width axes. For text tokens, both dimensions are set to $0$.
The sequential, temporal, height, and width dimensions each compute a separate RoPE, which are assigned RoPE embedding dimensions of $12$, $4$, $56$, $56$ respectively. 
To better support variable-length input, we use the NTK-aware interpolation~\citep{ntk} in RoPE calculation for context window extension.

\subsection{Training and Inference Paradigm} 
\label{sec:training_and_inference_paradigm}

Given an interleaved sequence $X_1 = \operatorname{Concat}(X^{(0)}_1 , X^{(1)}_1 , \ldots , X^{(n)}_1)$, where each $X^{(i)}_1$ represents a clean image, a video, or a text segment, and $n$ is the total number of visual or textual segments, we randomly select one image or video $X^{(i)}_1$ as the generation target, optimizing with the Flow Matching~\citep{flowmatching} training objective.
In the diffusion process with the formulation of Flow Matching, noise sample $X_0^{(i)} \sim \mathcal{N}(0,1)$ is progressively denoised into clean data $X_1^{(i)}$ with $X_t^{(i)} = tX_1^{(i)} + (1-t)X_0^{(i)}$, where timestep $t \in [0, 1]$. The learnable model $u$ is trained to predict the velocity $V_t=\frac{dX_t^{(i)}}{dt}$, which can be further derived as: $V_t=\frac{dX_t^{(i)}}{dt} = X_1^{(i)} - X_0^{(i)}$.
Thus, with an input sequence $X_t = \operatorname{Concat}(X^{(0)}_1 , \ldots , X^{(i)}_{t} , \ldots , X^{(n)}_1)$, the model $u$ with parameter $\Theta$ is optimized by minimizing the mean squared error loss $\mathcal{L}$ between the ground truth velocity and the  model prediction, where $X_0 = \operatorname{Concat}(X^{(0)}_1  , \ldots , X^{(i)}_{0} , \ldots , X^{(n)}_1)$:

$$\mathcal{L}=\mathbb{E}_{t,\mathbf{X}_0,\mathbf{X}_1}\left| u_{\Theta}(\mathbf{X}_t,t) - (\mathbf{X}_1 -  \mathbf{X}_0)  \right|^2
$$

During inference, the diffusion model first samples $X_0^{(i)} \sim \mathcal{N}(0,1)$, then uses an ODE solver with a discrete set of $N$ timesteps to generate $X_1$ from $X_0$.

\section{Data Pipeline}
\label{sec:dataset}

\OursMethod~is trained on large-scale data composed of:
$1.9M$ image generation samples (around $2.0B$ tokens), $3.9M$ video generation samples (around $68.8B$ tokens), $6.0M$ image editing samples (around $12.6B$ tokens), and $288K$ video editing samples (around $10.2B$ tokens).
Notably, the video editing datasets have significantly smaller sample number and are less diverse than image editing datasets. 
Thus, our architecture is specifically designed to transfer learned editing knowledge from the image domain to the video domain.
We summarize the used datasets in Table~\ref{tab:statistics_of_training_datasets} and provide further details below. A more detailed breakdown for each dataset is provided in Appendix Table~\ref{tab:detail_statistics_of_training_datasets}.

\begin{wraptable}{r}{0.45\textwidth}
\centering
\setlength\tabcolsep{5pt} 
\resizebox{0.45\textwidth}{!}{
\begin{threeparttable}
\begin{tabular}{ccc}
\toprule
\toprule
\multicolumn{3}{l}{\textbf{Image Datasets}}  \\ \midrule
& \textbf{Dataset}                 & \textbf{\#Samples}          \\ \midrule
 \multirow{10}{*}{\rotatebox{90}{\textbf{Edit}}}    & MagicBrush& $9K$     \\ 
         & ShareGPT-4o-Image& $46K$ \\
         & Object Removal \& Addition$^\ddag$ & $119K$ \\ 
         & OmniEdit& $186K$($1.2M^*$)    \\ 
         & ImgEdit& $246K$($1.2M^*$)   \\ 
         & NHR-Edit& $358K$ \\
         & UltraEdit& $500K$  \\ 
         & AnyEdit& $1.2M$($2.5M^*$)   \\ 
         & GPT-Image-Edit-1.5M& $1.5M$ \\
         & Instruction-based Editing$^\ddag$ & $1.8M$     \\ 
  \midrule
 \multirow{4}{*}{\rotatebox{90}{\textbf{Gen}}}         &   BLIP3-o 60K&  $60K$                             \\
         & LLaVA-pretrain& $500K$  \\
         & Text-to-Image$^\ddag$   &    $610K$       \\ 
         & LLaVA-next fine-tuning& $700K$\\
         \midrule \midrule
\multicolumn{3}{l}{\textbf{Video Datasets}}  \\ \midrule
& \textbf{Dataset}                 & \textbf{\#Samples}          \\ \midrule
 \multirow{2}{*}{\rotatebox{90}{\textbf{Edit}}} & Señorita-2M 
 & $56K$($2M^*$) \\
& \OursMethod~Editing Data & $232K$($1.3M^*$) \\
         \midrule
 \multirow{3}{*}{\rotatebox{90}{\textbf{Gen}}}  
  & Text-to-Video$^\ddag$  &  $223K$    \\
  & Customization  &  $740K$    \\
  & \OursMethod~Gen Data  &  $3.0M$    \\
 \bottomrule
 \bottomrule
 \multicolumn{3}{l}{\footnotesize $^*$Dataset volume before filtering. $^\ddag$ Internal dataset.}
\end{tabular}
\end{threeparttable}
}
\caption{\textbf{Statistics of the training datasets.} We mix open-source datasets, internal datasets, and \OursMethod~datasets for unified training. Detailed information in Table~\ref{tab:detail_statistics_of_training_datasets}.}
\label{tab:statistics_of_training_datasets}
\vspace{-5mm}
\end{wraptable}

\noindent \textbf{Video Editing Data Pipeline.} Due to the scarcity and inadequate quality of publicly available video editing datasets, we developed our pipeline to generate \textbf{\OursMethod~Editing Data}, which can be applied to obtain video editing pairs from any video input. (1) \textbf{Object Removal and Addition}. We first use Grounded-SAM-2~\citep{sam2,groundedsam} to extract object masks from the video. To improve the success rate of object removal, we filter candidates based on object name, total mask area, and detection confidence score. Next, we apply DiffuEraser~\citep{diffueraser} to remove the masked objects. We use video pairs before and after the removal to construct object removal and addition data. (2) \textbf{Object Change}. 
Again, we obtain object masks from Grounded-SAM-2~\citep{sam2,groundedsam}.
Next, we leverage a Vision-Language Model (VLM)~\citep{Qwen2VL} to imagine plausible transformations of the object. We then use VACE~\citep{vace} to inpaint the masked region based on the VLM's output. To improve the success rate, we apply dynamic adjustments to the mask’s shape and area, conditioned on the object's size and geometry. (3) \textbf{Style Transfer}. Previous style transfer techniques primarily rely on inference-based video editing methods~\citep{fatezero}, which we found to be unreliable when handling more diverse styles (\textit{e.g.}, Minecraft style). To address this, we first apply an image style transfer model to edit the first frame, and then utilize VACE’s~\citep{vace} depth-guided first-frame-to-video feature to generate the full styled video. (4) \textbf{Camera Change}. We select $10$ camera movements and use ReCamMaster~\citep{recammaster} to generate camera change data.  (5) \textbf{Mask Detection}. We construct the mask detection dataset by converting object removal, object addition, and object change data using the prompt template: ``I want to [edit prompt]. Detect the region that needs to be edited". (6) \textbf{Propagation}. We build the propagation dataset by extracting the first edited frame from style transfer, object removal, object addition, and object change data.

In addition, we incorporate data from the open-source dataset Señorita-2M~\citep{senorita-2m}. However, we observe a relatively low success rate in this dataset, necessitating extensive filtering. 

\noindent \textbf{Video Generation Data Pipeline.} 
Since we start from a pretrained model capable of text-to-image and text-to-video tasks, we only use a small scale of pure text-based generation data ($223K$ samples for text-to-video) to preserve the model's inherent generative capabilities while simultaneously introducing controllability and enhancing its text comprehension via control tasks. 
For controllable video generation, we annotate control-to-video and video-to-control data pairs (including depth, sketch, and pose), where the depth map is annotated with Depth Anything v2~\citep{depthanythingv2}, human pose is annotated with RTMPose~\citep{rtmpoose}, and sketch is annotated with OpenCV Canny Edge Detection~\citep{opencv}.
Moreover, we also include annotations for first-frame-to-video generation data and video inpainting data annotated with Grounded-SAM-2~\citep{sam2,groundedsam}.
The combined data from control-to-video, video-to-control, first-frame-to-video, and video inpainting are referred to as \textbf{\OursMethod~Gen Data} in Table~\ref{tab:statistics_of_training_datasets}.
Additionally, we include a video customization dataset to support reference-based generation~\citep{omnivcus}.

\noindent \textbf{Image Editing.} After reviewing the data quality of existing image editing datasets, we incorporate $8$ high-quality open-source datasets: MagicBrush~\citep{magicbrush}, ShareGPT-4o-Image~\citep{sharegpt4oimg}, OmniEdit~\citep{omniedit}, ImgEdit~\citep{imgedit}, NHR-Edit~\citep{nhredit}, UltraEdit~\citep{ultraedit}, AnyEdit~\citep{anyedit}, and GPT-Image-Edit-1.5M~\citep{gptimageedit}. 
In addition, we incorporate two internal image editing datasets: one focused on image addition and removal, and the other on free-form instruction-based image editing.

\noindent \textbf{Image Generation.} 
For text-to-image, we include $610K$ internal text-to-image samples as well as several open-source image understanding datasets (BLIP3-o 60K~\citep{blip3o}, LLaVA-pretrain~\citep{llava}, and LLaVA-next fine-tuning~\citep{llavanext}) that contain high-quality text annotations, which can improve the editing instructions understanding ability.

\noindent \textbf{Data Filtering.} Since the training data is model-generated and contains errors, filtering is vital for curating high-quality examples. We used a VLM~\citep{Qwen2VL} to filter the dataset by scoring both editing and video quality. The scores covered instruction adherence, context preservation, video sharpness, temporal consistency, artifact presence, object integrity, aesthetics, and physical plausibility. 
To determine the final filtering criteria, we manually inspected the relationship between the VLM scores and the editing quality. Based on this inspection, we defined a set of score thresholds to select the final training dataset. 
As shown in Table~\ref{tab:statistics_of_training_datasets}, our video editing pipeline achieves a retention rate six times higher than Señorita-2M after filtering, demonstrating high editing quality.

\vspace{-3mm}
\section{Experiments}
\label{sec:experiments}
\vspace{-3mm}

\subsection{Implementation Details}
\label{sec:implementation_details}
\vspace{-2mm}

\OursMethod~is trained on a $2B$ dense transformer architecture similar to LLaMA 3~\citep{llama3}. It is initially pretrained on text-to-image and text-to-video data to get basic generative capabilities at a resolution of 360p. Then, we train the model on our dataset as listed in Section~\ref{sec:dataset}. For each image/video, we resize it according to its original aspect ratio so that its area falls between 256×256 and 512×512. 
During training, we use a global batch size of $256$ and train for around $56K$ steps. 
We use AdamW optimizer~\citep{adamw} with hyper-parameters set to $\beta_{1}=0.9, \beta_{2}=0.95$, a peak learning rate of $8e^{-6}$, and weight decay of $0.01$.
We use a warm-up of $2K$ steps and a cosine decay learning schedule, decreasing the learning rate to the minimum of $1e^{-6}$.
We set the gradient clipping norm to 1.0 and disable gradient clipping during the warm-up stage.
Since the training data consist of token sequences with variable lengths, making it difficult to form batches, we adopt the packing strategy introduced in KnapFormer~\citep{knapformer}. During inference, we use a classifier-free guidance scale of $5.0$, applying it only to text conditions. The inference timestep is set to $50$ for the balance of performance and inference speed.

\begin{figure}[htbp]
    \centering
    \vspace{-2mm}
    \includegraphics[width=0.99\linewidth]{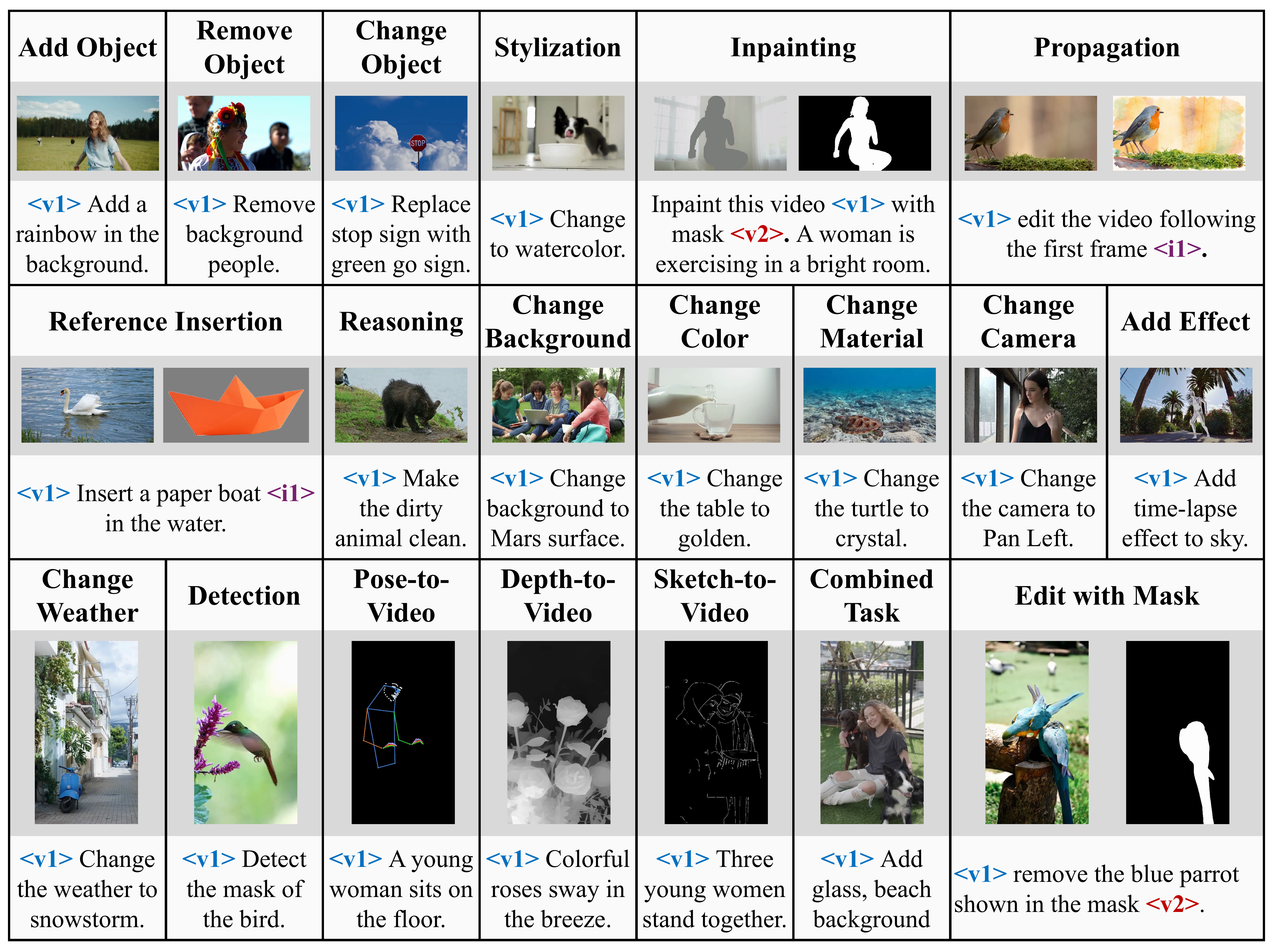}
    \vspace{-2mm}
    \caption{\textbf{Examples from the proposed \OursBench}. \OursBench~includes $200$ editing pairs, evenly distributed across 20 editing categories as well as horizontal and vertical orientations.
    } 
\label{fig:bench}
\end{figure}

\subsection{\OursBench}
\vspace{-2mm}

Commonly used video editing benchmarks (\textit{e.g.}, V2VBench~\citep{v2vbench}, TGVE~\citep{tgve,tgve+}) only consist of square videos and are primarily designed for training-free editing~\citep{tokenflow,stdf} rather than instruction-based editing. 
Moreover, such benchmarks do not adequately cover the diverse editing tasks commonly encountered in real-world video editing scenarios.
To address these limitations, we propose \OursBench, a comprehensive instruction-based video editing benchmark composed of $20$ distinct instruction-based video editing tasks. 
We manually selected $100$ videos from a free stock website~\citep{pixabay} that cover a variety of scenes, including $50$ horizontal and $50$ vertical videos. 
For each video, we randomly select two editing instructions from the 20 editing tasks. This results in a total of 200 editing pairs (5 horizontal and 5 vertical videos per editing task). We show one example from each editing category in Figure~\ref{fig:bench}. To evaluate editing performance on our proposed \OursBench, we use $6$ metrics covering four aspects: VLM evaluation, video quality (frame-wise Pick Score~\citep{pickscore}), text alignment (CLIP~\citep{clip} text-image and ViCLIP~\citep{internvid} text-video alignment), and temporal consistency (frame-wise CLIP~\citep{clip} and DINO~\citep{dino} consistency). Details can be found in the Appendix.

\vspace{-2mm}
\subsection{Comparison to Previous Methods}
\vspace{-2mm}

We show comparisons of \OursBench~and TGVE+~\citep{tgve+} in this section. More comparisons (\textit{e.g.}, V2VBench~\citep{v2vbench} and image editing) are provided in the Appendix.

\begin{table}[t]
\vspace{-2mm}
\centering
\resizebox{0.98\textwidth}{!}{
\begin{tabular}{l|c|c|cc|cc}
\toprule
\toprule
\multirow{2}{*}{\textbf{Method}} & \multicolumn{1}{c|}{\textbf{VLM evaluation}} & \multicolumn{1}{c|}{\textbf{Video Quality}} & \multicolumn{2}{c|}{\textbf{Text Alignment}}  &  \multicolumn{2}{c}{\textbf{Temporal Consistency}} \\ \cmidrule{2-7} 
 & \textbf{Editing Quality $\uparrow$}   & \textbf{Pick Score $\uparrow$}    & \textbf{Frame $\uparrow$}            & \textbf{Video $\uparrow$}      & \textbf{CLIP $\uparrow$}                & \textbf{DINO $\uparrow$}                                         \\ \midrule
\multicolumn{7}{c}{\textbf{Attention Manipulation (Training-free)}}  \\ 
\midrule
 \textbf{TokenFlow}         &        5.26     &      19.73          &          25.57        &        22.70        &       98.36              &       98.09                                     \\
 \textbf{STDF}  & 4.41 & 19.45  & 25.24 & 22.26     &    96.04 & 95.22   \\
\midrule
\multicolumn{7}{c}{\textbf{First-Frame Propagation (w/ End-to-End Training)}}  \\ 
\midrule
 \textbf{Señorita-2M} & 6.97 & 19.71 & 26.34 & 23.24    &  98.05 & 97.99      \\ 
\midrule
\multicolumn{7}{c}{\textbf{Instruction-Guided (w/ End-to-End Training)}}  \\ 
\midrule
 \textbf{InsV2V}        &         5.21   &         19.39   &     24.99             &       22.54  &       97.15              &      96.57                                                        \\
 \textbf{Lucy Edit}    & 5.89     & 19.67  & 26.00 & 23.11  &   98.49 & 98.38    \\
 \textbf{\OursMethod~(Ours)}  &    \textbf{7.65}         &   \textbf{20.07}   &     \textbf{26.73}             &    \textbf{23.93}     &       \textbf{98.56}              &       \textbf{98.42}                                                        \\ 
\midrule
 \midrule
\multicolumn{7}{c}{\textcolor{gray!130}{\textbf{Closed-Source Commercial Models}}}  \\ 
\graymidrule
   
 \multicolumn{1}{l!{\color{gray!130}\vrule}}{\textcolor{gray!130}{\textbf{Runway Aleph}}}
 &   \multicolumn{1}{c!{\color{gray!130}\vrule}}{\textcolor{gray!130}{7.44}}     &        \multicolumn{1}{c!{\color{gray!130}\vrule}}{\textcolor{gray!130}{20.42}}      &     \textcolor{gray!130}{27.70}             &      \multicolumn{1}{c!{\color{gray!130}\vrule}}{\textcolor{gray!130}{24.27}}   &      \textcolor{gray!130}{98.94}               &       \textcolor{gray!130}{98.60}                                                             \\

 \graybottomrule
 \graybottomrule
\end{tabular}
}
\vspace{-2mm}
\caption{\textbf{Quantitative comparison on \OursBench}. For open-source research models, we compare two training-free methods (TokenFlow and STDF), one first-frame propagation method (Señorita-2M), and one instruction-guided video editing method (InsV2V). Best results are highlighted in \textbf{bold}. We also provide the results of a commercial model, Runway Aleph. While \OursMethod~lags Runway Aleph in generation quality due to base model differences, our proposed method \OursMethod~surpasses it in editing faithfulness (via VLM evaluation on editing quality), aligning better with human judgment that is further validated by user studies shown in Figure~\ref{fig:user_study}.}
\label{tab:editverse_bench}
\vspace{-6mm}
\end{table}

\begin{wrapfigure}{r}{0.45\textwidth} 
\vspace{-4mm}
    \centering
    \includegraphics[width=0.45\textwidth]{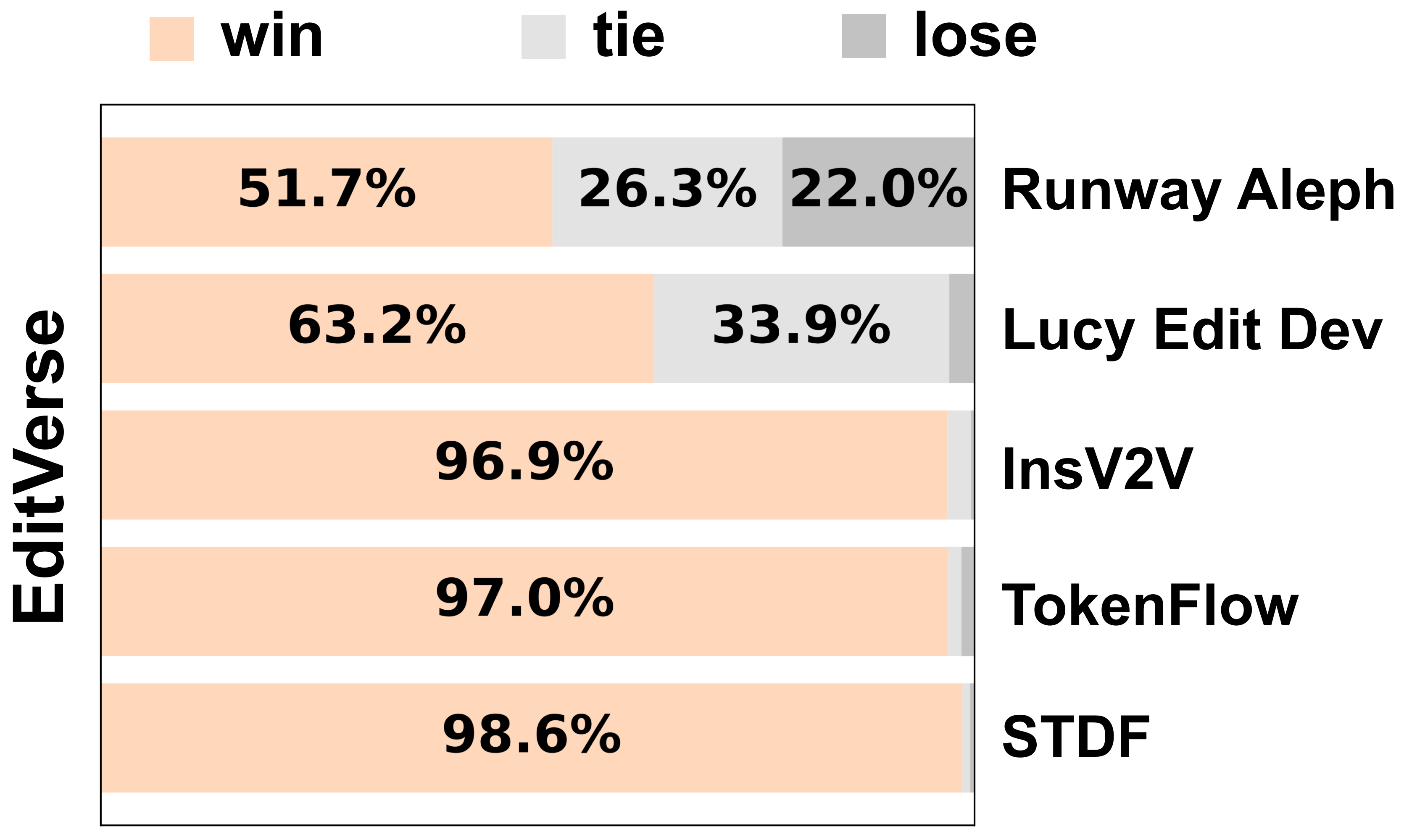}
\vspace{-4mm}
    \caption{\textbf{User study on \OursBench.}}
\vspace{-4mm}
    \label{fig:user_study}
\end{wrapfigure}

\begin{figure*}[htbp]
    \centering
\vspace{-2mm}
    \includegraphics[width=0.99\linewidth]{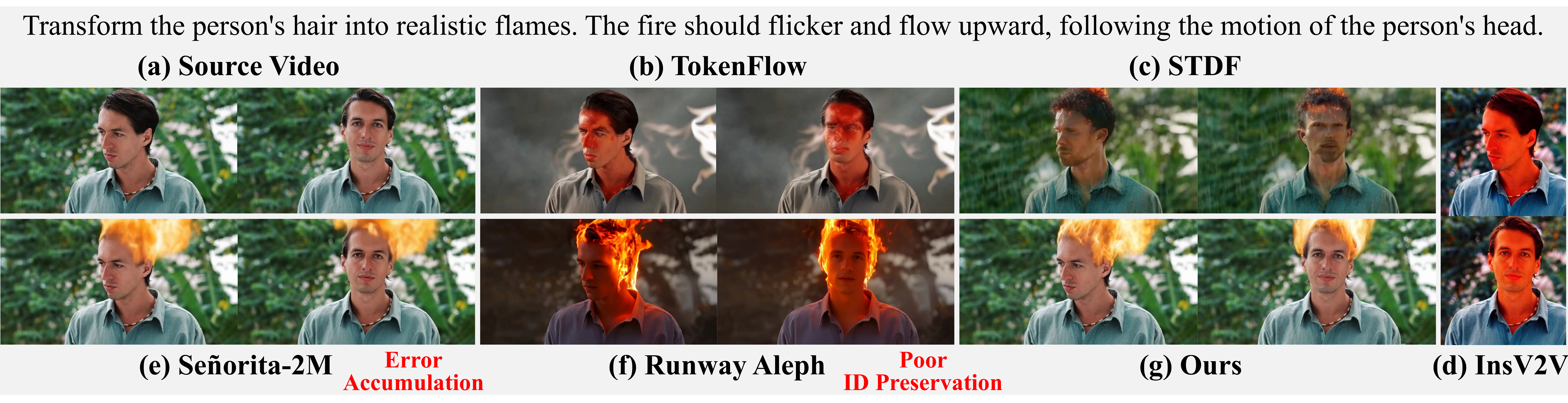}
\vspace{-2mm}
    \caption{\textbf{Visualization of \OursMethod~and other video editing methods}. EditVerse shows stronger context preservation and edit faithfulness. Complete comparisons are in the Appendix.
    } 
\label{fig:visual_compare}
\vspace{-2mm}
\end{figure*}

\noindent\textbf{Comparison on \OursBench}. Since InsV2V ~\citep{insv2v} and Lucy Edit~\citep{lucyedit} are the only open-source instruction-based video editing method that exactly matches our setting, we selected two well-known training-free methods, TokenFlow~\citep{tokenflow} and STDF~\citep{stdf}, as well as a first-frame propagation method, Señorita-2M~\citep{senorita-2m}, for comparison on \OursBench. 
We use the first frame of our results as input to Señorita-2M. 
Moreover, we also compare to a commercial model, Runway Aleph~\citep{runwayaleph}.
As shown in Table~\ref{tab:editverse_bench}, \OursMethod~outperforms previous research models on all metrics, demonstrating the effectiveness of our proposed method.
Figure~\ref{fig:visual_compare} shows visual comparisons on \OursBench. 
We further conduct a user study to assess human judgments of editing performance. The evaluation criteria include (i) instruction alignment, (ii) preservation of unedited regions, and (iii) overall video quality. 
We collected $3,000$ pairwise ratings comparing \OursMethod~against each of the other methods, with the results summarized in Figure~\ref{fig:user_study}, demonstrating the state-of-the-art performance of our proposed \OursMethod. We find the user study result is more aligned with the VLM evaluation metric in automatic evaluation.

\begin{wraptable}{r}{0.55\textwidth}
\vspace{-4mm}
\centering
\setlength\tabcolsep{3pt} 
\resizebox{0.55\textwidth}{!}{
\begin{threeparttable}
\begin{tabular}{lcccc}
\toprule
\textbf{Method} & \textbf{ViCLIP$_{dir}\uparrow$} & \textbf{ViCLIP$_{out}\uparrow$} \\ \midrule
\textbf{Tune-A-Video}~\citep{tuneavideo}                    & 0.131  & 0.242  \\
\textbf{TokenFlow}~\citep{tokenflow}              & 0.128 & 0.237  \\
\textbf{STDF}~\citep{stdf}                   & 0.093  & 0.227  \\
\textbf{Fairy}~\citep{fairy}        & 0.140  & 0.197  \\
\textbf{InsV2V}~\citep{insv2v}                 & 0.174  & 0.236  \\
\textbf{SDEdit}~\citep{sdedit}               & 0.131  & 0.241  \\
\textbf{EVE}~\citep{eve}                    & 0.198  & 0.251  \\
\textbf{Movie Gen Edit}~\citep{moviegen}              & \textbf{0.225}  & 0.248  \\ \midrule
\textbf{\OursMethod~(Ours)}                   & \textbf{0.225}  & \textbf{0.252}  \\ \bottomrule
\end{tabular}
\end{threeparttable}
}
\vspace{-2mm}
\caption{\textbf{Quantitative comparison on TGVE+}. Results show superior performance of \OursMethod.}
\label{tab:TGVE}
\vspace{-4mm}
\end{wraptable}

\noindent\textbf{Comparison on TGVE+}. Following Movie Gen~\citep{moviegen}, we evaluate  \OursMethod~on TGVE+~\citep{tgve+}. 
Specifically, we follow previous works and measure
(i) ViCLIP$_{dir}$: text-video direction similarity, which evaluates the alignment between changes in captions and corresponding changes in the videos, and (ii) ViCLIP$_{out}$: output similarity, which measures the
similarity between the edited video and the output caption. As shown in Table~\ref{tab:TGVE}, \OursMethod~surpasses previous methods on both metrics. It is worth noting that all TGVE+ videos are square, whereas our training data does not include any square video editing samples.

\vspace{-3mm}
\subsection{Analysis of Emergent Ability}
\vspace{-3mm}

Emergent ability is one of the most exciting phenomena observed in large-scale model training, arising as data and model capacity increase. In this section, we specifically analyze this phenomenon. 

\noindent \textbf{Demonstration of emergent ability}. We show the emergent ability of video editing in two aspects: (1) the model can perform editing tasks that were not present in the training data, and (2) for tasks included in the training data, the model’s performance can even surpass the ground-truth quality.

The video editing training data covers only a limited set of tasks, including camera changes, style transfer, mask detection, object modification (addition, removal, or replacement), and propagation. However, as shown in Figure~\ref{fig:teaser}, our model is capable of performing tasks beyond the training distribution (\textit{e.g.}, change material, change weather, add effects). Furthermore, it can also handle multiple tasks (\textit{e.g.}, reference insertion by combining customization with inpainting).

\begin{figure*}[h]
    \centering
\vspace{-4mm}
    \includegraphics[width=0.99
    \linewidth]{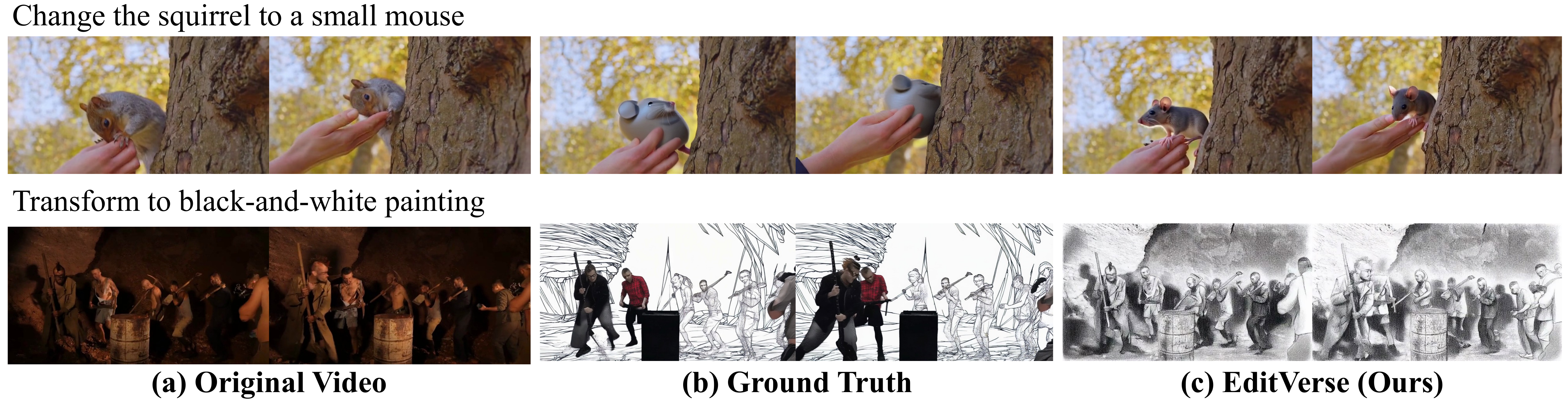}
\vspace{-2mm}
    \caption{\textbf{Compare \OursMethod~generated results with ground truth.} Results show \OursMethod~can surpass ground-truth data quality by extracting knowledge from image and video generation data.} 
\label{fig:compare_gt}
\end{figure*}

We also find that \OursMethod~can surpass the ground-truth training data in both quality and success rate by leveraging knowledge from the image generation/editing and video generation domains. We show two examples for object removal and object change in Figure~\ref{fig:compare_gt}.

\begin{figure*}[htbp]
\vspace{-2mm}
    \centering
    \includegraphics[width=0.99\linewidth]{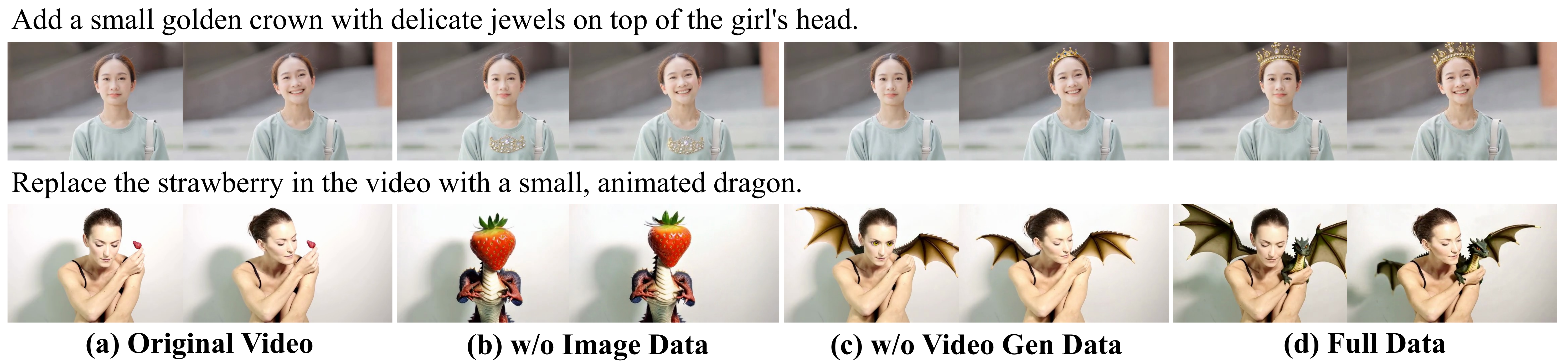}
\vspace{-2mm}
    \caption{\textbf{Visualization of ablation on training data.} Image data plays a critical role.
    } 
\label{fig:compare_w_full_data}
\vspace{-2mm}
\end{figure*}

\noindent \textbf{The source of emergent ability}. We further analyze the source of emergent ability by performing ablations on the training data. We find that removing either image generation/editing data or video generation data negatively impacts video editing quality. Specifically, image generation/editing data helps the model better understand editing instructions and perform more diverse edits, while video generation data improves temporal consistency and motion modeling. Figure~\ref{fig:compare_w_full_data} and Table~\ref{tab:image_videogen_dataset} illustrate the differences with and without image generation/editing and video data. Interestingly, \OursMethod~is able to perform some video editing tasks even without being trained on a video editing dataset.

\begin{table}[htbp]
\vspace{-1mm}
\centering
\setlength\tabcolsep{5pt} 
\resizebox{0.98\textwidth}{!}{
\begin{tabular}{ccc|c|c|cc|cc}
\toprule
\multicolumn{3}{c|}{\textbf{Training Datasets}} & \multicolumn{1}{c|}{\textbf{VLM evaluation}} & \multicolumn{1}{c|}{\textbf{Video Quality}} & \multicolumn{2}{c|}{\textbf{Text Alignment}}   & \multicolumn{2}{c}{\textbf{Temporal Consistency}} \\ \cmidrule{4-9} 
\textbf{Image} & \textbf{Video Gen} & \textbf{Video Edit}       & \textbf{Editing Quality}          & \textbf{Frame}            & \textbf{Video}            & \textbf{Pick Score}        & \textbf{CLIP}                & \textbf{DINO}   \\ \midrule
\ding{51}  &       \ding{51}    &  \ding{55} & 3.62 & 18.64           & 22.31 & 20.44   &   93.48    &   90.27 \\ 
\ding{55}  &       \ding{55}    &  \ding{51}  & 5.76      &     19.41                   &    25.22              &   22.37                              &    98.26                 &     97.83            \\ 
\ding{51}  &      \ding{55}   &  \ding{51}  & 6.52 & 19.81    & 25.78 & 22.63 &  98.24 & 97.97        \\
\ding{55}  &    \ding{51}   &  \ding{51}    &     6.40      &    19.72           &      25.37            &    22.51                      &      \textbf{98.77}               &     \textbf{98.60}               \\
\ding{51}  &    \ding{51}  &  \ding{51}     &    \textbf{6.95}   &      \textbf{19.99}           &    \textbf{26.26}              &     \textbf{23.81}                  &      98.68          &       98.44                     \\
 \bottomrule
\end{tabular}
}
\vspace{-3mm}
\caption{\textbf{Ablation study on training data}. We run $20K$ steps with the same setup as in Section~\ref{sec:implementation_details}. Results indicate that both image and video generation data are crucial to video editing performance.}
\label{tab:image_videogen_dataset}
\vspace{-3mm}
\end{table}

\vspace{-2mm}
\subsection{Ablation Study on Model Design}
\vspace{-2mm}

Compared with previous approaches~\citep{unireal}, our model contains two key designs: the interleaved formulation and the special positional embedding. Therefore, as shown in Table~\ref{tab:ablation}, we performed ablations by (i) removing the interleaved formulation (placing all images and videos at the end of the sequence) and (ii) removing the sequential dimension RoPE. 
Results show that both designs have a large influence on the model's performance, especially for the text alignment and editing quality. This is because the temporal consistency and video quality are partly inherited from the base model, while text alignment and editing quality largely depend on the in-context learning ability coming from the model design. Only the interleaved input format combined with sequential positional embedding can best enable the model to be aware of the relationships among different modalities (\textit{e.g.}, knowledge transfer of image and video), thereby achieving optimal performance.

\begin{table}[htbp]
\vspace{-2mm}
\centering
\resizebox{0.98\textwidth}{!}{
\begin{tabular}{cc|c|c|cc|cc}
\toprule
\multicolumn{2}{c|}{\textbf{Model Design}} & \multicolumn{1}{c|}{\textbf{VLM Evaluation}} & \multicolumn{1}{c|}{\textbf{Video Quality}} & \multicolumn{2}{c|}{\textbf{Text Alignment}}  & \multicolumn{2}{c}{\textbf{Temporal Consistency}} \\ \cmidrule{3-8} 
\textbf{Interleave} & \textbf{Sequential PE}    & \textbf{Editing Quality}     & \textbf{Pick Score}           & \textbf{Frame}            & \textbf{Video}                         & \textbf{CLIP}                & \textbf{DINO}      \\ \midrule
\ding{51}  &      \ding{55}       &       6.42   &        19.89                &       25.77         &     22.74                                &     98.62             &     98.43          \\
\ding{55}  &    \ding{51}            &      6.84          &     19.92           &        26.19            &   23.51                      &    \textbf{98.69}              &      98.39            \\
\ding{51}  &    \ding{51}      &   \textbf{6.95}   &    \textbf{19.99}                 &    \textbf{26.26}              &     \textbf{23.81}                     &        98.68          &       \textbf{98.44}         \\
 \bottomrule
\end{tabular}
}
\vspace{-2mm}
\caption{\textbf{Ablation study on interleaved formation and sequential RoPE}. We run $20K$ steps with the same experimental setting detailed in Section~\ref{sec:implementation_details} for the ablation to save compute.}
\vspace{-3mm}
\label{tab:ablation}
\end{table}

\vspace{-2mm}
\section{Conclusion}
\label{sec:conclusion}
\vspace{-2mm}

This paper introduced \OursMethod, a unified framework designed to address the architectural and data-scarcity challenges in universal video generation and editing. By representing text, images, and videos as a single interleaved token sequence, our model leverages full self-attention for robust in-context learning, enabling flexible inputs/outputs of arbitrary resolution and duration, while facilitating knowledge transfer from the data-abundant image domain to the video domain.

We further developed a data pipeline for obtaining high-quality video editing samples and proposed \OursBench, a benchmark covering diverse editing tasks. Results show that \OursMethod~achieves state-of-the-art performance. These findings validate that a unified architecture can mitigate video data limitations via cross-modal learning, revealing emergent abilities and paving the way for more general multimodal foundation models. Limitations and future work are discussed in the Appendix.

\newpage

\textbf{Acknowledgement.} 
This project was supported in part by the Innovation and Technology Fund (UD-1a-MHP/213/24), Hong Kong S.A.R.

\bibliography{iclr2026_conference}
\bibliographystyle{iclr2026_conference}

\newpage

\appendix
\section{Appendix}

\vspace{-2mm}
\subsection{Image and Video Copyrights}
\vspace{-2mm}

Figure~\ref{fig:teaser} videos are from $\mathtt{pixabay}$~\citep{pixabay}, stockbusters -- stock.adobe.com (the first video on the top), andreybiling -- stock.adobe.com (the second video on the top), and Mara Zemgaliete -- stock.adobe.com (the third video on the top). Comparison images in Figure~\ref{fig:teaser} are from ImgEdit-Bench~\citep{imgedit}. Example videos in Figure~\ref{fig:interleave} are from $\mathtt{pixabay}$~\citep{pixabay} and blackboxguild -- stock.adobe.com (the first video in ``More Examples"). Example videos in Figure~\ref{fig:bench},~\ref{fig:visual_compare}, and~\ref{fig:compare_w_full_data} are from $\mathtt{pixabay}$~\citep{pixabay}. Adobe Stock~\citep{adobe_stock} videos are officially licensed from the website. 

\vspace{-2mm}
\subsection{Evaluation Details}
\vspace{-2mm}

\noindent \textbf{Automatic Evaluation.} To provide a comprehensive and robust evaluation of instruction-based video editing models on \OursBench, we employ a suite of six metrics spanning four aspects: overall editing quality evaluated by a Vision-Language Model (VLM), video quality, text alignment, and temporal consistency. 

\vspace{-2mm}

\begin{itemize}
    \item \textbf{Overall Editing Quality Evaluated by VLM}: To evaluate the overall editing performance, we employ a state-of-the-art Vision-Language Model (VLM), GPT-4o~\cite{gpt-4o}, to serve as an automated judge. This provides a scalable, human-like assessment that captures nuances of editing quality, such as semantic correctness and artifact presence, which are often missed by other metrics. Our method uniformly samples three frames from each source and edited video pair. For each sample, the VLM receives the source frame, the edited frame, and the text instruction. It is prompted to score the edit from 0 (worst) to 3 (best) across three key criteria: Prompt Following, Edit Quality, and Background Consistency, and then sum them together to get the overall score for this frame. The final VLM score for the entire video is the average of these three frame scores.
    \item \textbf{Video Quality}: We employ PickScore~\citep{pickscore}, which shows a strong correlation with human judgment of image quality and prompt alignment. We calculate the PickScore for each frame and average these scores across the entire video.
    
    \item \textbf{Text Alignment}: Text alignment evaluates how well the edited video reflects the given text instruction. We measure this at both the frame level and the video level.

    \textit{CLIP Text-Image Alignment}: This metric assesses the semantic alignment between the editing instruction and each frame of the output video. We encode the text instruction using the CLIP text encoder and each frame using the CLIP vision encoder to get feature vectors, respectively. The final score is the average cosine similarity across all frames.

    \textit{ViCLIP Text-Video Alignment}: Frame-wise alignment doesn't capture the temporal aspects of the instruction. Therefore, we use ViCLIP~\citep{internvid} to compute an embedding for the entire video clip and measure its cosine similarity with the text instruction's embedding. This measures how well the video as a whole corresponds to the prompt.
    
    \item \textbf{Temporal Consistency}: Temporal consistency measures the smoothness and coherence of the edited video, penalizing flickering, jarring transitions, and inconsistent object appearances between frames. We assess this using feature similarity between adjacent frames.

    \textit{Frame-wise CLIP Consistency}: We use the ViT-L/14 vision encoder from CLIP~\citep{clip} to extract features of each frame in the edited video. The consistency score is calculated as the average cosine similarity between the features of all adjacent frames.

    \textit{Frame-wise DINO Consistency}: To capture more fine-grained structural and textural consistency, we repeat the same procedure using features extracted from a pre-trained DINOv2 model~\citep{dino}. DINO's self-supervised training allows it to capture object details that might be overlooked by CLIP. Similarly, the consistency score is calculated as the average cosine similarity between the features of all adjacent frames.
\end{itemize}

\vspace{-2mm}
\noindent \textbf{User Study.} To validate our automated metrics and directly measure human perceptual preferences, we conducted a comprehensive user study. 
The user study was outsourced to a professional external vendor, who recruited 20 annotators coming from diverse non-expert backgrounds. Each comparison pair was independently evaluated by 3 different annotators, and each annotator labeled 150 comparison pairs in total. To faithfully capture end-user preferences, we intentionally kept the instruction minimal and user-centric, providing only the brief prompt shown below, without additional technical guidance. This setup was chosen to reflect how typical users would judge the outputs rather than imposing task-specific expertise. Although we did not compute a formal inter-annotator agreement statistic, the redundancy of three independent judgments per pair helps mitigate noise and increases the robustness of the aggregated preferences. 
We recruited detail-oriented participants to evaluate the performance of different AI video editing models. Using a web-based interface, participants were shown pairs of edited videos, labeled ``Result 1" and ``Result 2", each generated by different models from the same source video and text instruction. Their task was to compare the two videos and choose among ``Result 1 is better," ``Result 2 is better," or ``They are about the same" across three evaluation criteria: (1) \textit{Text-Instruction Alignment}: Which video better follows the provided instruction? (2) \textit{Preservation of Unedited Regions}: Are unmodified parts of the video accurately preserved, with minimal distortion or artifacts? Ideally, edits should only affect the intended object or region. Select the one that preserves better. (3) \textit{Aesthetic Quality}: Which video is more visually appealing in terms of realism, smoothness, and overall perceptual quality? A video is considered the winner of a comparison if it achieves a majority of wins across these three criteria.
We find the user study shows a Pearson Correlation of 0.84 with automatic VLM evaluation, indicating a very strong positive correlation between the user study and VLM rankings.

\noindent \textbf{Instructions of Figure~\ref{fig:teaser}}. We list the editing instructions that were used in Figure~\ref{fig:teaser} in a top-to-bottom, left-to-right manner: (1) Add a pair of sparkling feathered wings to the person who is running. (2) Turn the man into a running cartoon leopard. (3) Turn the person into a translucent, crystal-glass-like form. (4) Remove the woman.  (5) Transform the woman’s dress into a golden, fluid-like form with flames.  (6) Turn into cartoon form. (7) Change the water to blue. (8) Change the camera pose to Pan Left. (9) Change the woman's slip dress to red and add a gentle snowfall effect. (10) Turn the grass into a reflective water surface. (11) Dramatically transform the scene by adding animated fiery embers and gentle flame wisps subtly dancing along the edges of the rose petals, giving the impression that the flower is being ignited by magical fire without harm, creating a surreal and striking contrast of beauty and intensity. (12) Insert a paper boat in the water [source image] A graceful white swan glides silently across the still surface of a clear lake, its long neck curved in a gentle arch and its feathers shining with a soft pearly sheen in the sunlight. Beside it, an orange paper boat drifts lightly, its sharp folds and pointed bow creating small ripples as it floats. (13) Two vibrant blue parrots are perched closely together on a tree stump. They appear to be pecking or searching for food in the crevice of the wood. The background shows a sunlit, green outdoor area with other birds visible in the distance, giving the scene a lively and natural atmosphere. (14) Change the weather to a heavy snowfall. (15) Detect the mask of the bird. (16) A young beautiful woman wearing a white hijab and a long white top sits quietly on the floor. She is reading from an open book, which rests on an intricately carved wooden stand. Her expression is calm and focused as she moves her finger along the lines of text, absorbed in her reading. The peaceful setting, with soft light and a tiled background, suggests a moment of reflection or prayer. (17) A quiet tree-lined path stretches into the distance, bathed in soft sunlight. Green leaves form a canopy overhead, while brown and yellow leaves are scattered across the ground. The scene feels calm and peaceful, inviting a slow walk or a moment of reflection in nature.

\vspace{-2mm}

\subsection{Additional Experiments}

\vspace{-2mm}

\noindent\textbf{Image Editing.} We present a comprehensive evaluation of \OursMethod~for the task of image editing using the ImgEdit-Bench benchmark, as summarized in Table~\ref{tab:imgedit_results}. The results demonstrate that \OursMethod~achieves highly competitive performance, surpassing a wide range of existing approaches~\citep{begal,step1xedit}. This highlights the effectiveness of our method.

\begin{table}[htbp]
\vspace{-2mm}
\resizebox{0.98\textwidth}{!}{
\begin{tabular}{l|ccccccccc|c}
\toprule
\textbf{Method}       & \textbf{Add} & \textbf{Adjust} & \textbf{Extract} & \textbf{Replace} & \textbf{Remove} & \textbf{Background} & \textbf{Style} & \textbf{Hybrid} & \textbf{Action} & \textbf{Overall$\uparrow$} \\ \midrule
\textbf{MagicBrush}   &  2.84   &    1.58    &   1.51      &    1.97     &    1.58    &    1.75        &   2.38    &     1.62   &  1.22       &   1.83      \\
\textbf{Instruct-P2P} &  2.45   &    1.83    &    1.44     &     2.01    &  1.50      &    1.44        &   3.55    &   1.20     & 1.46       &   1.88      \\
\textbf{AnyEdit} &  3.18   &    2.95    &    1.88     &     2.47    &  2.23      &    2.24        &   2.85    &   1.56     & 2.65       &   2.45      \\
\textbf{UltraEdit} &  3.44  &   2.81    &   2.13     &  2.96      &   1.45    &   2.83      &  3.76    &   1.91    &   2.98     &    2.70   \\
\textbf{ICEdit} &  3.58  &   3.39    &    1.73    &   3.15     &   2.93    &  3.08       &   3.84   &   2.04    &    3.68    &   3.05    \\
\textbf{Step1X-Edit} & 3.88   &   3.14    &  1.76      &     3.40   &  2.41     &   3.16      &  4.63    &   2.64    &     2.52   &   3.06    \\
\textbf{UniWorld-V1} &  3.82  &   3.64    &  2.27      &    3.47    &    3.24   &  2.99       &  4.21    &    2.96   &    2.74    &    3.26   \\
\textbf{BAGEL} &  3.81  &   3.59    &    1.58    &    3.85    &   3.16    &    3.39     &   4.51   &    2.67   &    4.25    &   3.42    \\
\rowcolor{gray!20}\textbf{\OursMethod~(Ours)} & 3.81 & 3.62 & 1.44 & 3.95 & 3.14 & 3.58 & 4.71 & 2.72 & 3.80 & 3.42  \\
\textbf{OmniGen2} &  3.57  &   3.06    &   1.77     &   3.74     &   3.20    &    3.57     &  4.81    &   2.52    &    4.68    &   3.44   \\
\textbf{Kontext-dev} &  3.83  &   3.65    &  2.27      &   4.45     &   3.17    &    3.98     &   4.55   &   3.35    &   4.29     &   3.71    \\
\textbf{Ovis-U1} & 3.99   &   3.73    &    2.66    &    4.38    &   4.15    &     4.05    &  4.86    &  3.43     &   4.68     &    3.97   \\
\textbf{GPT-4o-Image} &  4.61  &   4.33    &  2.90      &    4.35    &    3.66   &    4.57     &   4.93   &   3.96    &    4.89    &   4.20    \\
\bottomrule
\end{tabular}}
\vspace{-2mm}
\caption{\textbf{Quantitative comparison on ImgEdit-Bench}~\citep{imgedit}.}
\vspace{-2mm}
\label{tab:imgedit_results}
\end{table}

\noindent\textbf{Video Generation}. We evaluate the video generation capability of \OursMethod~on the VBench benchmark~\citep{vbench}, shown in Table~\ref{tab:vbench}. As shown, \OursMethod~achieves highly competitive performance compared with a wide range of both open-source and commercial models. Notably, even though \OursMethod~is trained on diverse tasks beyond video generation and is built with a relatively small model size, it can still match or surpass the performance of several larger-scale systems. 

\begin{table}[htbp]
    \centering
\resizebox{0.72\textwidth}{!}{
\begin{tabular}{lcccc}
\toprule
          
          Models &  \# Params.  & Total & Quality Score & Semantic Score  \\

\midrule
ModelScope & 1.7B & 75.75 & 78.05 & 66.54  \\
LaVie & 3B & 77.08 & 78.78 & 70.31  \\
OpenSoraPlan V1.3 & - & 77.23 & 80.14 & 65.62  \\
Show-1 & 6B & 78.93 & 80.42 & 72.98  \\
AnimateDiff-V2 & - & 80.27 & 82.90 & 69.75  \\
Gen-2 & - & 80.58 & 82.47 & 73.03 \\
Pika-1.0 & - & 80.69 & 82.92 & 71.77  \\
VideoCrafter-2.0 & - & 80.44 & 82.20 & 73.42  \\
\rowcolor{gray!20} EditVerse (Ours) & 2B & 80.97 & 83.47 & 70.97  \\
CogVideoX & 5B & 81.61 & 82.75 & 77.04 \\
Kling & - & 81.85 & 83.39 & 75.68  \\
Step-Video-T2V & 30B & 81.83 & 84.46 & 71.28 \\
Gen-3 & -& 82.32 & 84.11 & 75.17 \\ 
\bottomrule
\end{tabular}}
\caption{\textbf{Comparison with text-to-video models on the VBench}~\citep{vbench}. \# Params. is the number of total parameters. \OursMethod~shows competitive performance with a small model size.} 

\vspace{-2mm}
    \label{tab:vbench}
\end{table}

\noindent\textbf{Image Generation}. We evaluate the image generation capability of \OursMethod\ using the GenEval benchmark~\citep{geneval} shown in Table~\ref{tab:geneval}, which is designed to comprehensively assess text-to-image models across multiple aspects of visual reasoning and compositional fidelity. Our method achieves state-of-the-art performance when compared against a wide range of both open-source and commercial systems, highlighting better semantically aligned generation.

\begin{table}[htbp]
\vspace{-2mm}
\resizebox{0.98\textwidth}{!}{
\begin{threeparttable}
\begin{tabular}{l|cccccc|c}
\toprule
\textbf{Method}       & \textbf{Single Obj.} & \textbf{Two Obj.} & \textbf{Counting} & \textbf{Colors} & \textbf{Position} & \textbf{Color Attri.} & \textbf{Overall} \\ \midrule
\textbf{LlamaGen}     & 0.71        & 0.34     & 0.21     & 0.58   & 0.07     & 0.04         & 0.32    \\
\textbf{LDM}          & 0.92        & 0.29     & 0.23     & 0.70   & 0.02     & 0.05         & 0.37    \\
\textbf{SDv1.5}       & 0.97        & 0.38     & 0.35     & 0.76   & 0.04     & 0.06         & 0.43    \\
\textbf{PixArt-Alpha} & 0.98        & 0.50     & 0.44     & 0.80   & 0.08     & 0.07         & 0.48    \\
\textbf{SDv2.1}       & 0.98        & 0.51     & 0.44     & 0.85   & 0.07     & 0.17         & 0.50    \\
\textbf{DALL-E 2}     & 0.94        & 0.66     & 0.49     & 0.77   & 0.10     & 0.19         & 0.52    \\
\textbf{Emu3-Gen}     & 0.98        & 0.71     & 0.34     & 0.81   & 0.17     & 0.21         & 0.54    \\
\textbf{SDXL}         & 0.98        & 0.74     & 0.39     & 0.85   & 0.15     & 0.23         & 0.55    \\
\textbf{DALL-E 3}     & 0.96        & 0.87     & 0.47     & 0.83   & 0.43     & 0.45         & 0.67    \\
\textbf{Infinity$^\dag$}   & -        &  0.85     & -    & -   & 0.49     & 0.57         & 0.73   \\
\textbf{SD3-Medium}   & 0.99        & 0.94     & 0.72     & 0.89   & 0.33     & 0.60         & 0.74   \\
\textbf{FLUX.1-dev$^\dag$}   & 0.98        & 0.93     & 0.75     & 0.93   & 0.68     & 0.65         &  0.82   \\
\rowcolor{gray!20}\textbf{EditVerse (Ours)}   & 0.99        & 0.95     & 0.81     & 0.82   & 0.68     & 0.64         &  0.82   \\
\bottomrule
\end{tabular}
\begin{tablenotes} 
        \footnotesize 
        \item[$\ddag$] use LLM-rewritten prompts.
    \end{tablenotes}
\end{threeparttable}}
\caption{\textbf{Comparison with text-to-image models on the GenEval}~\citep{vbench}. } 

\vspace{-2mm}
\label{tab:geneval}
\end{table}

\noindent \textbf{Video Editing.} We provide a quantitative comparison on V2VBench~\citep{v2vbench} in Table~\ref{tab:v2vbench}. Noted that all V2VBench videos are square, whereas our training data does not include any square video editing samples. Our method achieves the best or competitive results across most metrics.

\begin{table}[htbp]
\centering
\setlength\tabcolsep{2pt} 
\resizebox{0.98\textwidth}{!}{
\begin{tabular}{lccccccc}
\toprule
\textbf{Method}          & \begin{tabular}[c]{@{}c@{}}\textbf{Frames}\\ \textbf{Quality} $\uparrow$\end{tabular} & \begin{tabular}[c]{@{}c@{}}\textbf{Semantic}\\ \textbf{Consistency} $\uparrow$\end{tabular} & \begin{tabular}[c]{@{}c@{}}\textbf{Object}\\ \textbf{Consistency} $\uparrow$\end{tabular} & \begin{tabular}[c]{@{}c@{}}\textbf{Frames Text}\\ \textbf{Alignment} $\uparrow$\end{tabular} & \begin{tabular}[c]{@{}c@{}}\textbf{Frames}\\ \textbf{Pick Score} $\uparrow$\end{tabular} & \begin{tabular}[c]{@{}c@{}}\textbf{Video Text}\\ \textbf{Alignment} $\uparrow$\end{tabular} & \begin{tabular}[c]{@{}c@{}}\textbf{Motion}\\ \textbf{Alignment} $\uparrow$\end{tabular} \\
\midrule
\multicolumn{8}{c}{Network and Training Paradigm}  \\
\midrule
\textbf{Tune-A-Video}    & \textbf{5.001}                                           & 0.934                                                          & 0.917                                                                                                     & 27.513                                                          & 20.701                                                      & 0.254                                                          & -5.599                                                     \\
\textbf{SimDA}           & 4.988                                                    & 0.940                                                          & 0.929                                                                                                     & 26.773                                                          & 20.512                                                      & 0.248                                                          & -4.756                                                     \\
\textbf{VidToMe}         & 4.988                                                    & 0.949                                                 & 0.945                                                                                             & 26.813                                                          & 20.546                                                      & 0.240                                                          & -3.203                                                     \\
\textbf{VideoComposer}   & 4.429                                                    & 0.914                                                          & 0.905                                                                                                       & 28.001                                                 & 20.272                                                      & 0.262                                                          & -8.095                                                     \\
\textbf{MotionDirector}  & 4.984                                                    & 0.940                                                          & 0.951                                                                                      & 27.845                                                          & 20.923                                             & 0.262                                                 & -3.088                                            \\ \midrule
\textbf{\OursMethod~(Ours)} & 4.957 & \underline{\textbf{0.959}} & \underline{\textbf{0.960}} & \textbf{28.587} &\underline{\textbf{21.117}} & \underline{\textbf{0.273}} & \textbf{-3.015}  \\
\midrule \arrayrulecolor{gray!100}
\multicolumn{8}{c}{\textcolor{gray!100}{Attention Feature Injection}}  \\
\midrule
\textcolor{gray!100}{\textbf{Video-P2P}}       & \textcolor{gray!100}{4.907}                                                    & \textcolor{gray!100}{0.943}                                                          & \textcolor{gray!100}{0.926}                                                                                          & \textcolor{gray!100}{23.550}                                                          & \textcolor{gray!100}{19.751}                                                      & \textcolor{gray!100}{0.193}                                                          & \textcolor{gray!100}{-5.974}                                                  \\
\textcolor{gray!100}{\textbf{Vid2Vid-Zero}}   & \textcolor{gray!100}{5.103}                                                    & \textcolor{gray!100}{0.919}                                                          & \textcolor{gray!100}{0.912}                                                                                                      & \textcolor{gray!100}{\textbf{28.789}}                                                 & \textcolor{gray!100}{\textbf{20.950}}                                             & \textcolor{gray!100}{\textbf{0.270}}                                                 & \textcolor{gray!100}{-4.175}                                                     \\
\textcolor{gray!100}{\textbf{Fate-Zero}}       & \textcolor{gray!100}{5.036}                                                    & \textcolor{gray!100}{\textbf{0.951}}                                                 & \textcolor{gray!100}{\textbf{0.952}}                                                                                            & \textcolor{gray!100}{25.065}                                                          & \textcolor{gray!100}{20.707}                                                      & \textcolor{gray!100}{0.225}                                                          & \textcolor{gray!100}{\textbf{-1.439}}                                           \\
\textcolor{gray!100}{\textbf{TokenFlow}}       & \textcolor{gray!100}{5.068}                                                    & \textcolor{gray!100}{0.947}                                                          & \textcolor{gray!100}{0.943}                                                                                            & \textcolor{gray!100}{27.522}                                                          & \textcolor{gray!100}{20.757}                                                      & \textcolor{gray!100}{0.254}                                                          & \textcolor{gray!100}{-1.572}                                                     \\
\textcolor{gray!100}{\textbf{FLATTEN}}         & \textcolor{gray!100}{4.965}                                                    & \textcolor{gray!100}{0.943}                                                          & \textcolor{gray!100}{0.949}                                                                                                     & \textcolor{gray!100}{27.156}                                                          & \textcolor{gray!100}{20.745}                                                      & \textcolor{gray!100}{0.251}                                                          & \textcolor{gray!100}{-1.446}                                                     \\
\textcolor{gray!100}{\textbf{FRESCO}}          & \textcolor{gray!100}{\underline{\textbf{5.127}}}                                           & \textcolor{gray!100}{0.908}                                                          & \textcolor{gray!100}{0.896}                                                                                                      & \textcolor{gray!100}{25.639}                                                          & \textcolor{gray!100}{20.239}                                                      & \textcolor{gray!100}{0.223}                                                          & \textcolor{gray!100}{-5.241}                                                     \\
\midrule
\multicolumn{8}{c}{\textcolor{gray!100}{Diffusion Latent Manipulation}}  \\
\midrule
\textcolor{gray!100}{\textbf{Text2Video-Zero}} & \textcolor{gray!100}{5.097}                                                    & \textcolor{gray!100}{0.899}                                                          & \textcolor{gray!100}{0.894}                                                                                                     & \textcolor{gray!100}{\underline{\textbf{29.124}}}                                                & \textcolor{gray!100}{20.568}                                                      & \textcolor{gray!100}{0.265}                                                          & \textcolor{gray!100}{-17.226}                                                    \\
\textcolor{gray!100}{\textbf{Pix2Video}}       & \textcolor{gray!100}{5.075}                                                    & \textcolor{gray!100}{0.946}                                                          & \textcolor{gray!100}{0.944}                                                                                                  & \textcolor{gray!100}{28.731}                                                          & \textcolor{gray!100}{\textbf{21.054}}                                            & \textcolor{gray!100}{\textbf{0.271}}                                               & \textcolor{gray!100}{-2.889}                                                     \\
\textcolor{gray!100}{\textbf{ControlVideo}}    & \textcolor{gray!100}{\textbf{5.404}}                                         & \textcolor{gray!100}{\textbf{0.959}}                                               & \textcolor{gray!100}{\textbf{0.948}}                                                                                         & \textcolor{gray!100}{28.551}                                                          & \textcolor{gray!100}{20.961}                                                      & \textcolor{gray!100}{0.261}                                                          & \textcolor{gray!100}{-9.396}                                                     \\
\textcolor{gray!100}{\textbf{Rerender}}        & \textcolor{gray!100}{5.002}                                                    & \textcolor{gray!100}{0.872}                                                          & \textcolor{gray!100}{0.863}                                                                                                   & \textcolor{gray!100}{27.379}                                                          & \textcolor{gray!100}{20.460}                                                      & \textcolor{gray!100}{0.261}                                                          & \textcolor{gray!100}{-4.959}                                                     \\
\textcolor{gray!100}{\textbf{RAVE}}            & \textcolor{gray!100}{5.077}                                                    & \textcolor{gray!100}{0.926}                                                          & \textcolor{gray!100}{0.936}                                                                                                  & \textcolor{gray!100}{28.190}                                                          & \textcolor{gray!100}{20.865}                                                      & \textcolor{gray!100}{0.255}                                                          & \textcolor{gray!100}{\textbf{-2.398}}  \\ \arrayrulecolor{black}

\bottomrule
\end{tabular}}
\caption{\textbf{Quantitative comparison on V2VBench~\citep{v2vbench}.} Methods are grouped into three categories: (i) Network and Training Paradigm, (ii) Attention Feature Injection, and (iii) Diffusion Latent Manipulation. Local best are in \textbf{bold}. Global best are \underline{underlined}.}
\vspace{-5mm}
\label{tab:v2vbench}
\end{table}

\vspace{-2mm}
\subsection{Detailed Training Data}
\vspace{-2mm}

Table~\ref{tab:detail_statistics_of_training_datasets} provides a detailed statistics overview of the whole training datasets that are used in our work, along with their respective ratio in the training process. The table is organized by task type, image editing, image generation, video editing, and video generation. For each dataset, we report the total number of samples, the ratio applied when constructing the training mixture, and a brief description highlighting the data quality, coverage, and characteristics. The training data comprises a mixture of high-quality open-source data, curated internal datasets, and filtered synthetic datasets. This combination allows us to balance scale, quality, and diversity, ultimately supporting unified training across both editing and generation tasks for images and videos.

\begin{table}[htbp]
\resizebox{0.95\textwidth}{!}{
\begin{threeparttable}
\begin{tabular}{crrc}
\toprule
\textbf{Dataset}                          & \textbf{\#Samples}     & \textbf{\#Ratio}          & \textbf{Information}                     \\ \midrule
\rowcolor{gray!10}\multicolumn{4}{c}{\textbf{Image Editing}}                                                                       \\ \midrule
\textbf{MagicBrush} & 8,802 & 10 &     \begin{tabular}[c]{@{}c@{}}Manually annotated with real image.\\ High-quality. $7$ editing categories.\end{tabular}\\ \graymidrule
\textbf{ShareGPT-4o-Image} & 46,489 & 10 & \begin{tabular}[c]{@{}c@{}}Generated by GPT-4o. $14$ editing categories.\\Most are high-quality, but some cases contain noise. \end{tabular} \\ \graymidrule
\textbf{Object Removal \& Addition$^\ddag$} & 118,972 & 4 & \begin{tabular}[c]{@{}c@{}}Manually captured photos with object-present \\and object-absent scenes. High-quaity.\end{tabular} \\ \graymidrule
\textbf{OmniEdit$^*$} & 185,500 & 2 & \begin{tabular}[c]{@{}c@{}}Generated by task-specific models. $7$ editing categories. \\Good-quality but contains large noise in some editing categories.\end{tabular} \\ \graymidrule
\textbf{ImgEdit$^*$} & 245,986 & 1 & \begin{tabular}[c]{@{}c@{}}Generated by segmentation and inpainting. $13$ editing categories. \\Fair quality. Need filtering.\end{tabular} \\ \graymidrule
\textbf{NHR-Edit}  & 358,463 & 5 & \begin{tabular}[c]{@{}c@{}}Generated with a designed pipeline using internal \\image editing model. High-quality. $17$ editing categories.\end{tabular} \\ \graymidrule
\textbf{UltraEdit}  & 500,000 & 1 & \begin{tabular}[c]{@{}c@{}}Generated by a specially designed editing model. \\ Fair quality. $9$ editing categories.\end{tabular} \\ \graymidrule
\textbf{AnyEdit$^*$}  & 1,244,033 & 1 & \begin{tabular}[c]{@{}c@{}}Generated by task-specific pipelines. $25$ editing categories.\\Fair quality. Need filtering.\end{tabular} \\ \graymidrule
\textbf{GPT-Image-Edit-1.5M}  & 1,500,000  & 1 & \begin{tabular}[c]{@{}c@{}}Re-process OmniEdit, UltraEdit, and HQ-Edit with GPT-4o.\\
Most are high-quality, but some cases contain noise.\end{tabular} \\ \graymidrule
\textbf{Instruction-based Editing$^\ddag$} & 1,824,969 & 1 & \begin{tabular}[c]{@{}c@{}}An internal instruction-based image editing dataset.\end{tabular} \\ \midrule
\textbf{Sum} & \textbf{6,033,214} \\

\midrule
\rowcolor{gray!10}\multicolumn{4}{c}{\textbf{Image Generation}}                                                                       \\ \midrule
\textbf{BLIP3o-60k} & 60,000 & 1 & \begin{tabular}[c]{@{}c@{}}Text-to-Image instruction tuning dataset distilled from GPT-4o.\end{tabular} \\ \graymidrule
\textbf{LLaVA-pretrain} & 500,000 & 1 & \begin{tabular}[c]{@{}c@{}}Text-to-Image data re-captioned using Qwen2-VL (from text-to-image-2M).\end{tabular} \\ \graymidrule
\textbf{Text-to-Image$^\ddag$} & 609,950 & 1 & \begin{tabular}[c]{@{}c@{}}Internal high-quality text-to-image dataset.\end{tabular} \\  \graymidrule
\textbf{LLaVA-next fine-tuning} & 700,000 & 1 & \begin{tabular}[c]{@{}c@{}}Text-to-Image data generated by Flux-dev (from text-to-image-2M).\end{tabular} \\ \midrule
\textbf{Sum} & \textbf{1,869,950} \\

\midrule
\rowcolor{gray!10}\multicolumn{4}{c}{\textbf{Video Editing}}                                                                       \\ 
\midrule
\textbf{Camera Change} & 8,000 & 20 & \begin{tabular}[c]{@{}c@{}}Camera change data pair generated with ReCamMaster\end{tabular} \\ \graymidrule
\textbf{Style Transfer} & 10,327 & 10 & \begin{tabular}[c]{@{}c@{}} Style transfer data pair generated with Step1X-Edit and VACE.\end{tabular} \\ \graymidrule
\textbf{Mask Detection} & 15,741 & 5 & \begin{tabular}[c]{@{}c@{}} Editing region detection with prompt \\``I want to [edit prompt]. Detect the region that needs to be edited".\\Contain object removal, object addition, and object replacement.\end{tabular} \\ \graymidrule
\textbf{Object Chnage} & 31,482 & 10 & \begin{tabular}[c]{@{}c@{}}Object replacement data pair generated with VACE.\\Contain w/ mask version and w/o mask version in training.\end{tabular} \\ \graymidrule
\textbf{CG Removal \& Addition$^\ddag$} & 38,900 & 2 & \begin{tabular}[c]{@{}c@{}}Rendered videos with object-present and object-absent scenes.\end{tabular} \\ \graymidrule
\textbf{Señorita-2M$^*$} & 55,711 & 2 & \begin{tabular}[c]{@{}c@{}}Generated wih task-specific models. 5 editing categories. \\Low quality. Need filtering.\end{tabular} \\ \graymidrule
\textbf{Propagation} & 59,826 & 10 & \begin{tabular}[c]{@{}c@{}}Containing editing propagation for object removal, object addition, \\object replacement, and style transfer.\end{tabular} \\ \graymidrule
\textbf{Object Removal \& Addition} & 67,516 & 10 & \begin{tabular}[c]{@{}c@{}}Object removal and addition pairs generated with DiffuEraser.\\Contain w/ mask version and w/o mask version in training.\end{tabular} \\ \midrule
\textbf{Sum} & \textbf{287,503} \\

\midrule
\rowcolor{gray!10}\multicolumn{4}{c}{\textbf{Video Generation}}                                                                       \\ \midrule
\textbf{Depth-to-Video} & 182,097 & 2 & \begin{tabular}[c]{@{}c@{}} Depth-to-video dataset. Depth is detected with Depth Anything v2.\end{tabular} \\ \graymidrule
\textbf{Video-to-Depth} & 182,097 & 2 & \begin{tabular}[c]{@{}c@{}} Video-to-depth dataset. Depth is detected with Depth Anything v2.\end{tabular} \\ \graymidrule
\textbf{Sketch-to-Video} & 207,749 & 2 & \begin{tabular}[c]{@{}c@{}} Sketch-to-video dataset. Sketch is detected with OpenCV Canny. \end{tabular} \\ \graymidrule
\textbf{Video-to-Sketch} & 207,749 & 2 & \begin{tabular}[c]{@{}c@{}} Video-to-sketch dataset. Sketch is detected with OpenCV Canny. \end{tabular} \\ \graymidrule
\textbf{First Frame-to-Video} & 217,038 & 5 & \begin{tabular}[c]{@{}c@{}} First frame-to-video dataset.\end{tabular} \\ \graymidrule
\textbf{Pose-to-Video} & 233,068 & 2 & \begin{tabular}[c]{@{}c@{}} Pose-to-video dataset. Pose is detected with RTM-Pose.\end{tabular} \\ \graymidrule
\textbf{Video-to-Pose} & 233,068 & 2 & \begin{tabular}[c]{@{}c@{}} Video-to-pose dataset. Pose is detected with RTM-Pose.\end{tabular} \\ \graymidrule
\textbf{Text-to-Video$^\ddag$} & 223,494 & 10 & \begin{tabular}[c]{@{}c@{}}Internal high-quality text-to-video dataset.\end{tabular} \\ \graymidrule
\textbf{Customization} & 740,111 & 1 & \begin{tabular}[c]{@{}c@{}}High-quality identity-to-video dataset from OmniVCus~\citep{omnivcus}.\end{tabular} \\ \graymidrule
\textbf{Video Inpainting} & 1,495,020 & 2 & \begin{tabular}[c]{@{}c@{}}Video inpainting data pair generated with Grounded SAM 2.\\Contain w/ mask version and w/o mask version in training.\end{tabular} \\ \midrule
\textbf{Sum} & \textbf{3,921,491} \\
\bottomrule
\end{tabular}
\begin{tablenotes} 
        \footnotesize 
        \item[$\ddag$] Internal datasets.
        \item[$*$] We filter these datasets to improve their quality.
    \end{tablenotes}
\end{threeparttable}}
\caption{\textbf{Detailed Statistics of the training datasets.} We combine high-quality open-source datasets, internal datasets, and \OursMethod~datasets for unified training. This table presents the dataset name, sample counts, training ratios, and key details for each dataset.}
\label{tab:detail_statistics_of_training_datasets}
\end{table}

\vspace{-2mm}
\subsection{Limitation and Future Work}
\vspace{-2mm}

While \OursMethod~presents a significant step toward unified video and image generation and editing, we acknowledge several limitations that open avenues for future research.

\begin{figure*}[tbph]
\vspace{-2mm}
    \centering
    \includegraphics[width=0.99\linewidth]{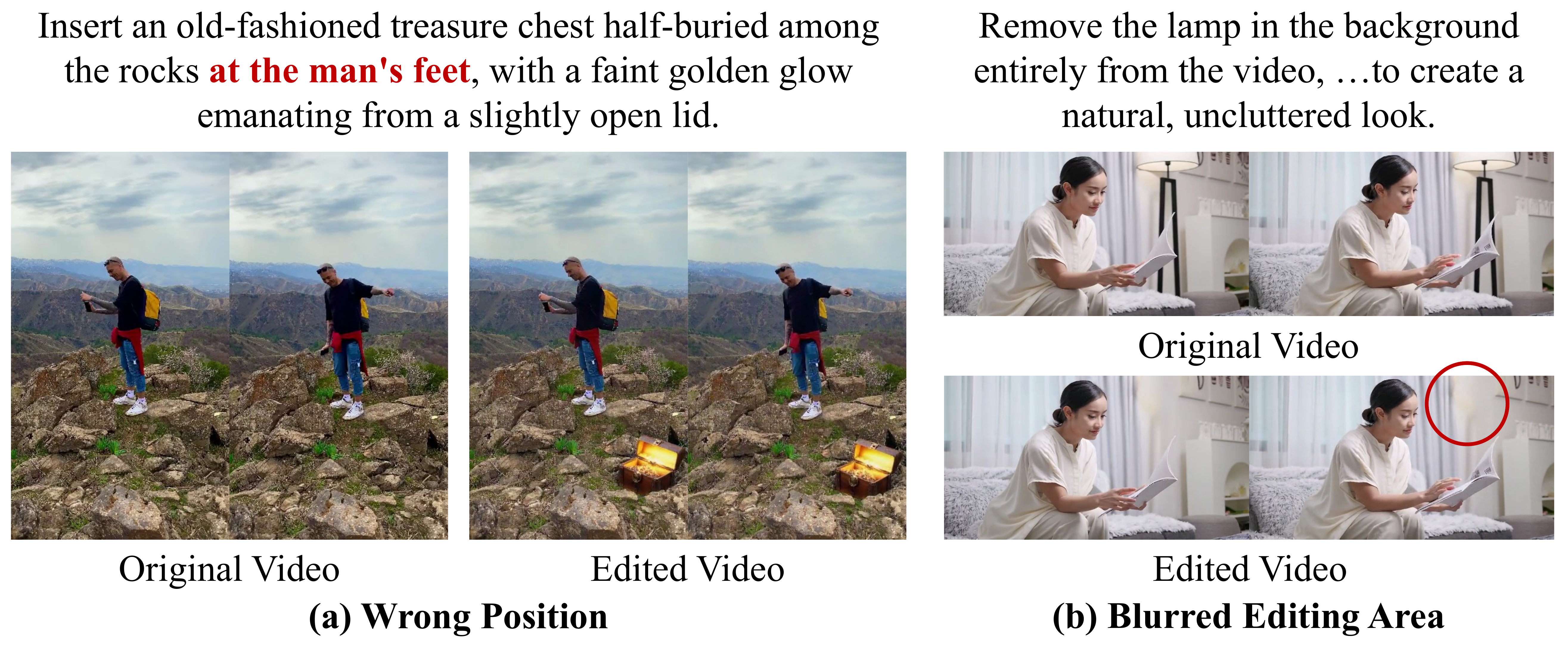}
\vspace{-3mm}
    \caption{\textbf{Failure case examples of \OursMethod}. (a) The model fails to add object (treasure chest) at the correct position (at the man's feet). (b) Generation of blurry artifacts within the edited region.
    } 
\vspace{-3mm}
\label{fig:fail}
\end{figure*}

\textbf{Observed Failure Cases.} Despite its strong overall performance, \OursMethod~is not immune to failure cases including artifacts, flickering, low motion, logical flaws, wrong editing position, and blurred editing region. Figure~\ref{fig:fail} shows examples of two commonly seen failure types of \OursMethod.

\textbf{Computational Cost.} Our reliance on a full self-attention mechanism across a unified one-dimensional token sequence, while powerful for in-context learning, leads to significant computational overhead. The concatenation operation results in long sequence lengths, particularly for high-resolution or long-duration videos, which translates to high FLOPs and prolonged training and inference time. Future work could explore more efficient attention mechanisms to reduce the computational burden without compromising the model's cross-modal learning capabilities.
Our work was not specifically designed to improve efficiency, and the reported memory usage and inference time reflect the model's raw, unoptimized performance. There are several practical ways to improve it, which we plan to explore in future work: (1) Using a higher-compression VAE. In our model, we use a VAE with relatively low compression ($8\times$ spatial downsampling), which leads to a large number of visual tokens. Recent VAEs can achieve $16\times$ spatial compression. If we replace our VAE with $16\times$ spatial compression rate models, the token length can be reduced to about one quarter, which directly lowers the attention cost. (2) Dynamic token selection. We can introduce a dynamic token selection mechanism that adaptively keeps only the most important context tokens and prunes redundant ones. This can reduce the effective sequence length for full attention, as explored in recent work FullDiT2~\citep{fulldit2}. (3) Distillation for faster inference. We can further apply model distillation and step distillation to reduce both the number of diffusion steps and the model size, which can noticeably speed up generation. (4) Future work could explore more efficient attention mechanisms (\textit{e.g.}, linear attention, Mamba attention) to reduce the computational burden without compromising the model's cross-modal learning capabilities.

\begin{figure}[htbp]
\centering
\vspace{-3mm}
    \includegraphics[width=0.4\textwidth]{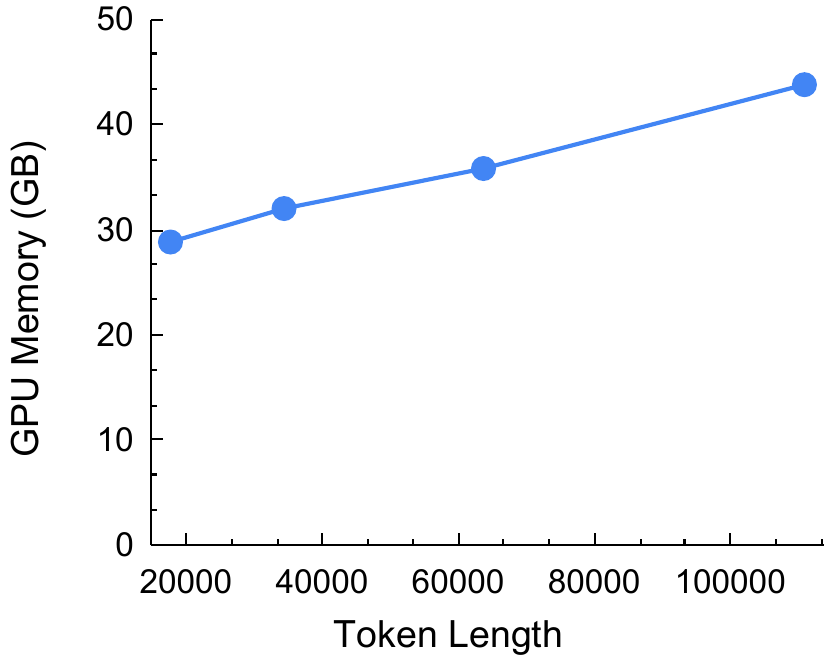}
    \label{fig:memory}
\hspace{6mm}
    \includegraphics[width=0.38\textwidth]{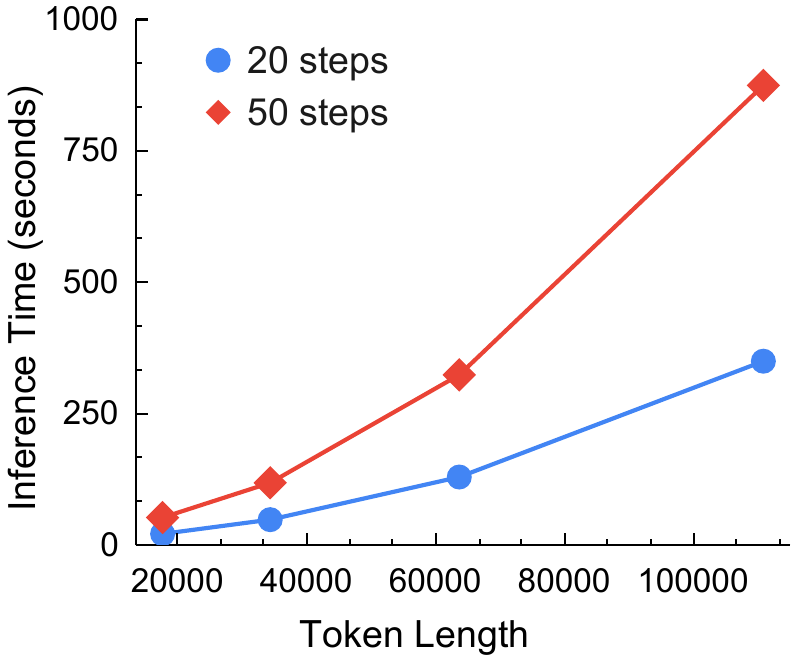}
    \label{fig:time}
\vspace{-2mm}
\caption{GPU memory usage and inference latency as functions of token length.}
\vspace{-2mm}
\label{fig:time_memory}
\end{figure}

\begin{table}[ht]
\centering
\scalebox{0.85}{
\begin{tabular}{l|rrrr}
\toprule
\textbf{Token Length}
  & 17652
  & 34342
  & 63662
  & 110822 \\
\midrule
\textbf{Inference Time (second/20 steps)}
  & 21  
  & 48
  & 129
  & 349 \\
\textbf{Inference Time (second/50 steps)} 
  & 52  
  & 118 
  & 323 
  & 873 \\
\textbf{GPU Memory (MB)}            
  & 29511 
  & 32768 
  & 36663 
  & 44824 \\
\textbf{TFLOPs/second}
  & 67.9
  & 58.2
  & 39.4
  & 25.4 \\
\bottomrule
\end{tabular}}
\vspace{-2mm}
\caption{Inference efficiency across different token lengths}
\label{tab:eff}
\vspace{-2mm}
\end{table}

To assess the efficiency of our model with extended token length (as shown in Figure~\ref{fig:time_memory} and Table~\ref{tab:eff}), we analyze the impact of token length on both efficiency and GPU memory usage. Experiment results demonstrate that resource consumption growth remains within a manageable and predictable range. Specifically, GPU Memory usage exhibits a strong linear relationship with token length. As the sequence length was scaled from 17,652 up to 110,822 tokens, the peak memory footprint increased modestly from 29.5 GB to 44.8 GB. This predictable and relatively slow scaling rate confirms that the extended context itself does not impose an unconstrained memory ceiling. Similarly, inference time shows a systematic, controllable increase with token length.

\textbf{Image Editing Performance.} While our unified model demonstrates strong generalization and performs on par with many image editing models, it does not currently achieve state-of-the-art performance in the image domain. Targeted optimizations, such as employing a more sophisticated data-mixing strategy or fine-tuning the model on high-quality, image-only editing datasets, could be explored to boost its performance and close the gap with specialized, state-of-the-art image editors.

\textbf{Dataset Quality.} Although our data curation pipeline is crucial for enabling instruction-based video editing, the resulting dataset contains inherent noise. The editing instructions are often concise (averaging around 10 words) and may lack the detail required for highly complex or nuanced edits. Furthermore, the automated methods used to generate editing pairs have an estimated success rate of around 65\%, inevitably introducing imperfect or failed edits into the training corpus. Future efforts could focus on developing more advanced data generation and filtering techniques as mentioned in concurrent works~\citep{wei2025univideo,mou2025instructx,bai2025scaling,yang2025unified}.

\textbf{Generalist vs. Specialist Models.} Our work highlights the potential of unified models, but it is plausible that for specific, well-defined tasks with abundant high-quality data (e.g., inpainting), a dedicated specialist model might still yield superior results. 

\end{document}